%% file: neurips_2024.tex
\documentclass{article}

\usepackage[nonatbib, final]{neurips_2024}

\usepackage[utf8]{inputenc} 
\usepackage[T1]{fontenc}    
\usepackage{url}            
\usepackage{booktabs}       
\usepackage{amsfonts}       
\usepackage{nicefrac}       
\usepackage{microtype}      
\usepackage{xcolor}         
\usepackage{makecell} 

\usepackage{graphicx}
\usepackage{enumitem}
\usepackage{tikz}
\usetikzlibrary{shapes.geometric, arrows, positioning}
\usepackage{cite}
\usepackage{forest}
\usetikzlibrary{shadows}
\usepackage{graphicx}
\usepackage{enumitem}
\usepackage{tabularx}
\usepackage{wrapfig}
\usepackage{multirow}

\newcolumntype{Y}{>{\setlength\hsize{2\hsize}}X}

\definecolor{hidden-red}{RGB}{205, 44, 36}

\usepackage[pagebackref=true,breaklinks=true,letterpaper=true,colorlinks,bookmarks=false,citecolor=hidden-red,linkcolor=hidden-red, urlcolor=hidden-red]{hyperref}

\usepackage{tcolorbox}
\newcommand{\insightbox}[1]{%
    \begin{tcolorbox}[colframe=black!60, colback=blue!5, boxrule=1pt, arc=4mm]
        \includegraphics[width=0.4cm]{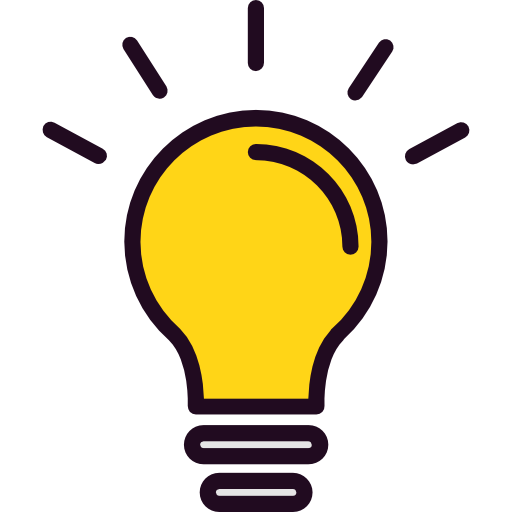}
        \textbf{\textit{#1}}
    \end{tcolorbox}
}

\usepackage{amsmath}

\title{Stop Overthinking: A Survey on \\ Efficient Reasoning for Large Language Models}

\author{%
\textbf{Yang Sui}$^{1}$ \enspace
\textbf{Yu-Neng Chuang}$^{1}$ \enspace
\textbf{Guanchu Wang}$^{1}$ \enspace
\textbf{Jiamu Zhang}$^{1}$  \enspace
\textbf{Tianyi Zhang}$^{1}$ \enspace
\textbf{Jiayi Yuan}$^{1}$ \\
\textbf{Hongyi Liu}$^{1}$ \enspace
\textbf{Andrew Wen}$^{1}$ \enspace
\textbf{Shaochen Zhong}$^{1}$ \enspace
\textbf{Na Zou}$^{2}$ \enspace
\textbf{Hanjie Chen}$^{1}$ \enspace
\textbf{Xia Hu}$^{1}$ \enspace \\ 
$^{1}$Rice University \enspace \enspace $^{2}$University of Houston \\
\texttt{yang.sui@rice.edu, xia.hu@rice.edu}
\\
\\
Project Website: {\href{https://github.com/Eclipsess/Awesome-Efficient-Reasoning-LLMs}{https://github.com/Eclipsess/Awesome-Efficient-Reasoning-LLMs}}
}

\begin{document}

\maketitle
\renewcommand{\thefootnote}{\fnsymbol{footnote}}
\renewcommand*{\thefootnote}{\arabic{footnote}}

\begin{abstract}
    Large Language Models (LLMs) have demonstrated remarkable capabilities in complex tasks. Recent advancements in Large Reasoning Models (LRMs), such as OpenAI o1 and DeepSeek-R1, have further improved performance in System-2 reasoning domains like mathematics and programming by harnessing supervised fine-tuning (SFT) and reinforcement learning (RL) techniques to enhance the Chain-of-Thought (CoT) reasoning. 
    However, while longer CoT reasoning sequences improve performance, they also introduce significant computational overhead due to verbose and redundant outputs, known as the ``overthinking phenomenon''.

    \textbf{\textit{Efficient Reasoning}}, which seeks to optimize reasoning length while preserving reasoning capabilities, offers practical benefits such as reduced computational costs and improved responsiveness for real-world applications. 
    Despite its potential, efficient reasoning remains in the early stages of research. 
    In this paper, we provide the first structured survey to systematically investigate and explore the current progress toward achieving efficient reasoning in LLMs. 
    Overall, relying on the inherent mechanism of LLMs, we categorize existing works into several key directions: \textbf{\textit{(1) model-based efficient reasoning}}, which considers optimizing full-length reasoning models into more concise reasoning models or directly training efficient reasoning models; \textbf{\textit{(2) reasoning output-based efficient reasoning}}, which aims to dynamically reduce reasoning steps and length during inference; \textbf{\textit{(3) input prompts-based efficient reasoning}}, which seeks to enhance reasoning efficiency based on input prompt properties such as difficulty or length control. 
    Additionally, we introduce the use of efficient data for training reasoning models, explore the reasoning capabilities of small language models, and discuss evaluation methods and benchmarking.
    We maintain a public repository to continuously track and update the latest research in this promising area.

\end{abstract}

\begin{figure}[h]
    \centering
    \includegraphics[width=\linewidth]{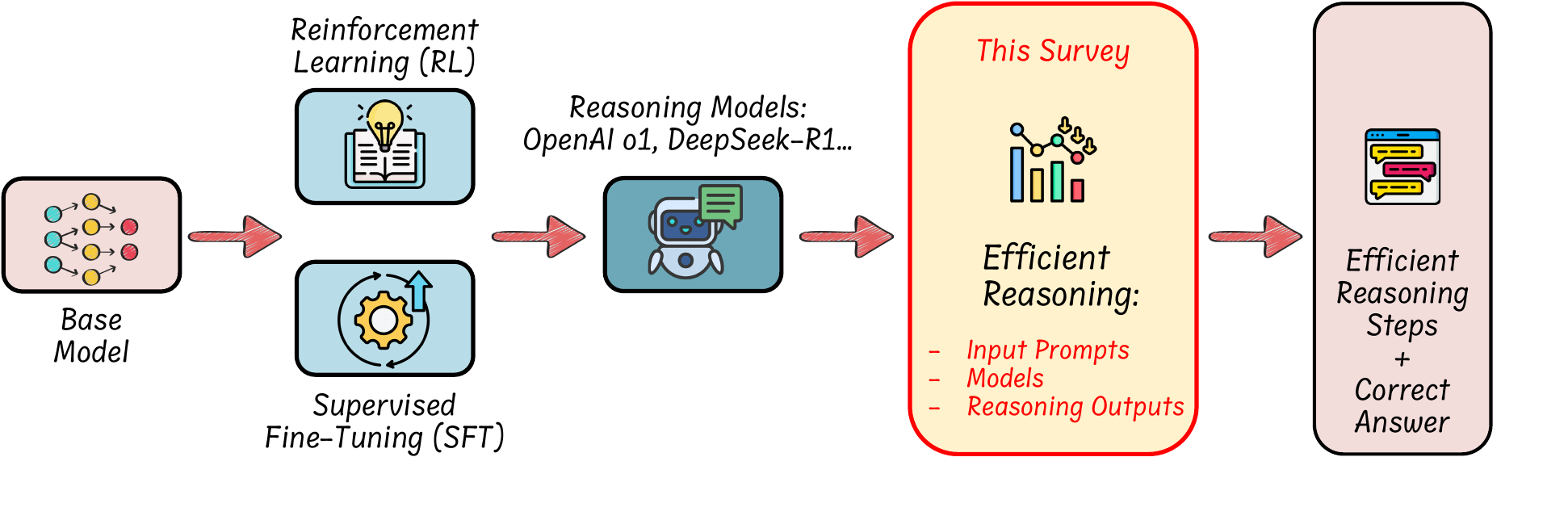}
    \caption{The pipeline of developing efficient reasoning for LLMs. A reasoning model can be trained on the base model using SFT, RL, or a combination of both. While reasoning models demonstrate strong reasoning capabilities, they often suffer from the ``overthinking phenomenon'', generating unnecessarily lengthy reasoning steps. To improve efficiency, various methods can be applied to reduce redundant steps while maintaining accuracy, or to fine-tune non-reasoning models to incorporate efficient reasoning capabilities. This approach enables the model to answer questions with concise and effective reasoning steps. In this paper, we explore the latest progress in efficient reasoning for LLMs, aiming to provide insights that can guide future research and the development of reasoning-driven applications across various domains.}
    \label{fig:pipeline}
\end{figure}

\input{sec/1_introduction}

\begin{figure}[t]
    \centering
    \includegraphics[width=\linewidth]{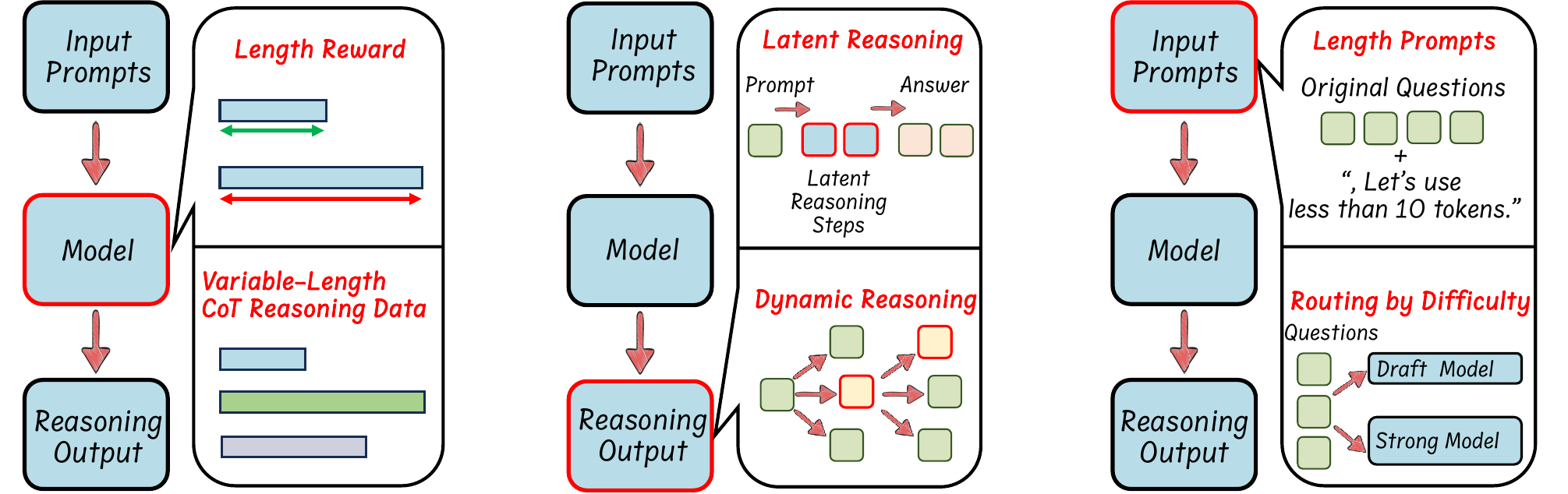}
    \vspace{2mm}
    \caption{
    Overview of efficient reasoning methods, which can be summarized as model-oriented (Left: \textit{I}, \textit{II}) and reasoning output-oriented (Middle: \textit{III}, \textit{IV}), and input prompts-oriented (Right: \textit{V, VI}) methods. Specifically, (I) Reinforcement Learning with Length Reward Design (Section \ref{sec:rl}); (II) Supervised Fine-Tuning with Variable-Length CoT Data (Section \ref{sec:longshortdata}); (III) Compressing Reasoning Steps into Fewer Latent Representation (Section \ref{sec:latent}); (IV) Dynamic Reasoning Paradigm during Inference (Section \ref{sec:dynamic}); (V) Prompt-guided Efficient Reasoning (Section \ref{sec:prompts}); (VI) Routing Prompts to Optimize Reasoning Efficiency (Section \ref{sec:routing}); }

    \label{fig:overview}
\end{figure}

\input{plot/taxonomy}
\input{sec/2_background}
\input{sec/part1_RL}

\input{sec/part2_data}
\input{sec/part3_latent}
\input{sec/part4_dynamic}

\input{sec/part5_prompt}

\input{sec/part6_lessdata}

\input{sec/part8_smallmodel}

\input{sec/part7_evaluation}
\input{sec/part11_application}
\input{sec/5_conclusion}


{
\bibliographystyle{plain}
\bibliography{ref}
}

\end{document}

%% file: sec/1_introduction.tex
\section{Introduction}

Large Language Models (LLMs) have emerged as exceptionally powerful AI tools, demonstrating advanced capabilities in natural language understanding and complex reasoning. Recently, the rise of reasoning-focused LLMs, also referred to as reasoning-capable models or Large Reasoning Models (LRMs) \cite{xu2025towards} such as OpenAI o1 \cite{openai_learning_to_reason} and DeepSeek-R1 \cite{guo2025deepseek}, has significantly improved performance in System-2 reasoning domains \cite{li2025system}, particularly in challenging mathematics \cite{cobbe2021training,hendrycks2measuring} and programming tasks \cite{codeforces, chen2021evaluating}. Evolving from foundational pretrained models (e.g., LLaMA \cite{touvron2023llama,grattafiori2024llama}) trained with next-token prediction \cite{devlin2019bert}, these models typically leverage Chain-of-Thought (CoT) \cite{wei2022chain} reasoning chains to generate explicit, step-by-step reasoning sequences before arriving at a final answer, significantly improving their effectiveness in reasoning-intensive tasks.

Such reasoning abilities in LLMs are typically developed through supervised fine-tuning (SFT) and reinforcement learning (RL), which promote iterative and systematic problem-solving abilities. For instance, DeepSeek-R1 \cite{guo2025deepseek} undergoes multiple rounds of SFT and RL training, emphasizing structured thinking templates and rule-based reward mechanisms. In particular, the rule-based rewards provides precise and explicit feedback signals during training, effectively enhancing the general reasoning capabilities beyond the pretrained LLM.

However, while long CoT reasoning significantly boosts accuracy, step-by-step thinking mechanisms also lead to lengthy output responses, resulting in substantial computational overhead and increased reasoning time. For instance, the "overthinking problem" arises when answering a simple question~\cite{chen2024not} like, "\textit{what is the answer of 2 plus 3?}" Some reasoning models, especially smaller ones, can generate reasoning sequences spanning thousands of tokens. This verbosity significantly increases both inference costs and latency, limiting the practical application of reasoning models in computation-sensitive real-world scenarios, such as real-time autonomous driving systems, interactive conversational assistants, precision robotic control tasks, and large-scale online search engines.

Efficient reasoning, particularly the reduction of reasoning length, offers significant benefits in such regards, providing direct cost reduction and improved feasibility for real-world deployments. Recently, numerous studies \cite{luo2025o1, yeo2025demystifying, han2024token, ma2025cot, hao2024training} have explored ways to develop more concise reasoning paths, making efficient reasoning a rapidly evolving research area.

In this paper, we present the first structured survey systematically exploring the progress in efficient reasoning for LLMs. As illustrated in Figure \ref{fig:overview}, we categorize existing work into three key directions:  
\textit{(1) Model-based efficient reasoning}, which focuses on optimizing full-length reasoning models into more concise variants or directly training efficient reasoning models.  
\textit{(2) Reasoning output-based efficient reasoning}, which dynamically reduces reasoning steps and length during inference.  
\textit{(3) Input prompts-based efficient reasoning}, which enhances reasoning efficiency based on input properties such as difficulty or length control. 
Unlike model compression techniques such as quantization \cite{xiao2023smoothquant, frantar2023gptq, lin2024awq} or KV cache compression \cite{zhang2023h2o, liu2024kivi, shi2024keepcost, yuan2024lcbench, hao2025omnikv}, which focus on reducing model size for lightweight inference, efficient reasoning in LLMs emphasizes \textit{smart and concise reasoning} by optimizing the length of \textit{generated} reasoning sequences and reducing unnecessary thinking steps.

Overall, we provide a summary of the current key approaches to efficient reasoning, organizing them into the following categories:

\begin{itemize}
    \item Reinforcement Learning with Length-Based Reward Design (Section \ref{sec:rl})
    \item Supervised Fine-Tuning with Variable-Length CoT Data (Section \ref{sec:longshortdata})
    \item Compressing Reasoning Steps into Fewer Latent Representations (Section \ref{sec:latent})
    \item Dynamic Reasoning Paradigms During Inference (Section \ref{sec:dynamic})
    \item Prompt-Guided Efficient Reasoning (Section \ref{sec:prompts})
    \item Routing Prompts to Optimize Reasoning Efficiency (Section \ref{sec:routing})
\end{itemize}

Additionally, we explore other relevant topics, including:

\begin{itemize}
    \item Training Reasoning Models with Efficient Data (Section \ref{sec:lessdata})
    \item Reasoning Abilities of Small Language Models and Model Compression (Section \ref{sec:compression}) 
    \item Evaluation and Benchmarking of Efficient Reasoning Models (Section \ref{sec:evaluation})
\end{itemize}

%% file: plot/taxonomy.tex
\label{taxonomy}

\tikzset{
    my node/.style={ 
        draw=black,
        fill=blue!1, 
        thick,
        rounded corners=5pt,
        text width=3.0cm, 
        align=left, 
        font=\bfseries, 
        drop shadow,
        inner sep=5pt, 
        minimum height=0.8cm, 
    },
    ERT node/.style={ 
        draw=black,
        fill=yellow!5, 
        thick,
        rounded corners=5pt,
        text width=3.0cm, 
        align=left, 
        font=\bfseries, 
        drop shadow,
        inner sep=5pt, 
        minimum height=0.8cm, 
    },
    ERI node/.style={ 
        draw=black,
        fill=blue!3, 
        thick,
        rounded corners=5pt,
        text width=3.0cm, 
        align=left, 
        font=\bfseries, 
        drop shadow,
        inner sep=5pt, 
        minimum height=0.8cm, 
    },
    OER node/.style={ 
        draw=black,
        fill=teal!10, 
        thick,
        rounded corners=5pt,
        text width=3.0cm, 
        align=left, 
        font=\bfseries, 
        drop shadow,
        inner sep=5pt, 
        minimum height=0.8cm, 
    },
    CP node/.style={ 
        draw=black,
        fill=red!7, 
        thick,
        rounded corners=5pt,
        text width=3.0cm, 
        align=left, 
        font=\bfseries, 
        drop shadow,
        inner sep=5pt, 
        minimum height=0.8cm, 
    },
    BI node/.style={ 
        draw=black,
        fill=green!5, 
        thick,
        rounded corners=5pt,
        text width=3.0cm, 
        align=left, 
        font=\bfseries, 
        drop shadow,
        inner sep=5pt, 
        minimum height=0.8cm, 
    },
    leaf node/.style={ 
        draw=black,
        fill=gray!5, 
        thick,
        rounded corners=5pt,
        text width=15cm, 
        align=left, 
        font=\bfseries, 
        drop shadow,
        inner sep=5pt,
        minimum height=1.6cm, 
    },
    ERTleaf node/.style={ 
        draw=black,
        fill=yellow!5, 
        thick,
        rounded corners=10pt,
        text width=15cm, 
        align=left, 
        font=\bfseries, 
        drop shadow,
        inner sep=5pt,
        minimum height=1.6cm, 
    },
    ERIleaf node/.style={ 
        draw=black,
        fill=blue!3, 
        thick,
        rounded corners=10pt,
        text width=15cm, 
        align=left, 
        font=\bfseries, 
        drop shadow,
        inner sep=5pt,
        minimum height=1.6cm, 
    },
    CPleaf node/.style={ 
        draw=black,
        fill=red!7, 
        thick,
        rounded corners=10pt,
        text width=15cm, 
        align=left, 
        font=\bfseries, 
        drop shadow,
        inner sep=5pt,
        minimum height=1.6cm, 
    },
    BIleaf node/.style={ 
        draw=black,
        fill=green!5, 
        thick,
        rounded corners=10pt,
        text width=15cm, 
        align=left, 
        font=\bfseries, 
        drop shadow,
        inner sep=5pt,
        minimum height=1.6cm, 
    },
    OERleaf node/.style={ 
        draw=black,
        fill=teal!10, 
        thick,
        rounded corners=10pt,
        text width=15cm, 
        align=left, 
        font=\bfseries, 
        drop shadow,
        inner sep=5pt,
        minimum height=1.6cm, 
    },
    taxonomy node/.style={ 
        draw=black,
        fill=gray!3, 
        thick,
        font=\bfseries, 
        text width=0.8cm, 
        align=center, 
        rotate=0, 
        inner sep=4pt,
        minimum height=2.0cm,
    },
    subtaxonomy node/.style={ 
        draw=black,
        fill=white!20, 
        thick,
        font=\bfseries, 
        text width=0.8cm, 
        align=center, 
        rotate=0, 
        inner sep=4pt,
        minimum height=2.0cm,
    },
    level 1/.style={edge={thick, black}},
    level 2/.style={edge={thick, black!70}},
    level 3/.style={edge={thick, black!50}},
}

\begin{figure}[t]
    \centering
    \resizebox{\textwidth}{!}{ 
        \begin{forest}
            for tree={%
                grow'=east,
                edge={thick},
                parent anchor=east,
                child anchor=west,
                l sep+=15pt, 
                s sep+=25pt, 
                if n children=0{leaf node}{my node}, 
                edge path={
                    \noexpand\path [draw, \forestoption{edge}] (!u.parent anchor) -- +(12pt,0) -| (.child anchor)\forestoption{edge label};
                },
                if={isodd(n_children())}{
                    for children={
                        if={equal(n,(n_children("!u")+1)/2)}{calign with current}{}
                    }
                }{}
            }
            [\rotatebox{90}{Taxonomy}, taxonomy node 
                [\parbox{3cm}{\centering Model-based \\ Efficient Reasoning}, ERT node
                    [\parbox{3cm}{\centering RL Optimization via Length Reward}, ERT node
                        [\parbox{15cm}{\raggedright
                        e.g. Kimi k1.5~\cite{team2025kimi}; O1-Pruner~\cite{luo2025o1}; L1~\cite{aggarwal2025l1}; Training~\cite{arora2025training}; Demystifying~\cite{yeo2025demystifying}; DAST~\cite{shen2025dast}; MRT~\cite{qu2025optimizing}; Self-adaptive~\cite{yang2025thinkneedselfadaptivechainofthought}; HAWKEYE~\cite{she2025hawkeyeefficientreasoningmodelcollaboration}; ThinkPrune~\cite{hou2025thinkprune}; LongShort~\cite{ning2025not}; ConciseRL~\cite{dumitru2025conciserl}; Bingo~\cite{liu2025bingo}; Concise Reasoning~\cite{fatemi2025concise}; Elastic Reasoning~\cite{xu2025scalable}; S-GRPO~\cite{dai2025s}; TLDR~\cite{zhang2025making}; SelfBudgeter~\cite{li2025selfbudgeter}; Short-RL~\cite{yuan2025efficient}; BRPO~\cite{qi2025optimizing}; LASER~\cite{liu2025learn}; ACPO~\cite{cheng2025incentivizing}; LIMOPro~\cite{xiao2025limopro}; L-GRPO~\cite{song2025walk}; GRPO-$\lambda$~\cite{dai2025stablereinforcementlearningefficient}; AutoThink~\cite{tu2025learningthinkshapingadaptive}; AdaptThink~\cite{zhang2025adaptthinkreasoningmodelslearn}; DeGRPO~\cite{fang2025thinklessllmlearnsthink}; HGPO~\cite{jiang2025thinkneedlargehybridreasoning}; DTO~\cite{an2025dontthinklongerthink}; REO-RL~\cite{gao2025faroptimalreasoningefficiency}; ALP~\cite{xiang2025justthinkingefficientreasoning}; PLP~\cite{ling2025fasteasydeephard}; LC-R1~\cite{cheng2025optimizinglengthcompressionlarge}; AdapThink~\cite{wan2025adapthinkadaptivethinkingpreferences}; AALC~\cite{li2025aalclargelanguagemodel}; DuP-PO~\cite{ding2025thinkingtokenshelptrap}; SCPO~\cite{he2025smartthinkerlearningcompresspreserve}; FCS~\cite{hong2025reconsideringoverthinkingpenalizinginternal}; CurriculumGRPO~\cite{hammoud2025trainlongthinkshort}; GFPO~\cite{shrivastava2025sample}; SABER~\cite{zhao2025saberswitchablebalancedtraining}; VSRM~\cite{yue2025promotingefficientreasoningverifiable}; DR.SAF~\cite{chen2025awarefirstthinkless}; ASRR~\cite{zhang2025continuethinkingadaptivethinking}; AdaCoT~\cite{lou2025adacotparetooptimaladaptivechainofthought}; 
                        \\ 
                        },  
                        ERTleaf node]
                    ]
                    [\parbox{3cm}{\centering SFT with Variable-Length CoT}, ERT node
                        [\parbox{15cm}{\raggedright
                        e.g. Distilling 2-1~\cite{yu2024distilling}; C3oT~\cite{kang2024c3ot}; TokenSkip~\cite{xia2025tokenskip}; CoT-Valve~\cite{ma2025cot}; Self-Training~\cite{munkhbat2025self}; Learn to Skip~\cite{liu2024can}; Token-Budget~\cite{han2024token}; Verbosity~\cite{jang2025verbosityawarerationalereductioneffective}; Stepwise~\cite{cui2025stepwise}; Z1~\cite{yu2025z1efficienttesttimescaling}; Prune-on-Logic~\cite{zhao2025can}; LS-Mixture SFT~\cite{yu2025long}; DRP~\cite{jiang2025drp}; AutoL2S~\cite{luo2025autol2s}; Assembly of Experts~\cite{klagges2025assembly}; Ada-R1~\cite{luo2025ada}; ConCISE~\cite{qiao2025concise}; VeriThinker~\cite{chen2025verithinker}; R1-Compress~\cite{wang2025r1}; CTS~\cite{yuan2025not}; A$^*$-Thought~\cite{xu2025athoughtefficientreasoningbidirectional}; TLDR~\cite{li2025tldrlongreweightingefficient}; OThink-R1~\cite{zhang2025othinkr1intrinsicfastslowthinking}; PNS~\cite{yu2025causalsufficiencynecessityimproves}; ReCUT~\cite{jin2025recutbalancingreasoninglength}; StepEntropy~\cite{li2025compressingchainofthoughtllmsstep}; ASAP~\cite{zeng2025pruningunsurprisingefficientcode}; 
                        \\
                        },
                        ERTleaf node]
                    ]
                ]
                [\rotatebox{0}{\parbox{3cm}{\centering Reasoning Output-based \\ Efficient Reasoning}}, OER node
                    [\parbox{3cm}{\centering Latent Representation Compression}, OER node
                        [\parbox{15cm}{\raggedright
                        e.g. Coconut~\cite{hao2024training}; CODI~\cite{shen2025codi}; CCoT~\cite{cheng2024compressed}; Heima~\cite{shen2025efficient}; Token Assorted~\cite{su2025token}; Loop~\cite{saunshi2025reasoning}; SoftCoT~\cite{xu2025softcot}; 
                        Back Attention~\cite{yu2025back}; CoLaR~\cite{tan2025think}; SEAL~\cite{chen2025seal}; Overclocking~\cite{eisenstadt2025overclockingllmreasoningmonitoring}; Controlling~\cite{lin2025controllingthinkingspeedreasoning}; 
                        \\
                        },
                        OERleaf node]
                    ]
                    [\parbox{3cm}{\centering Dynamic Reasoning Paradigm}, OER node
                        [\parbox{15cm}{\raggedright
                        e.g. Speculative Rejection~\cite{sun2024fast}; Sampling-Efficient TTS~\cite{wang2025sampling}; DPTS~\cite{ding2025dynamic}; Certaindex~\cite{fu2024efficiently}; Dynasor-CoT~\cite{fu2025reasoning}; Fast MCTS~\cite{li2025fastmcts}; ST-BoN~\cite{wang2025sampling}; More is Less~\cite{wu2025more}; RSD \cite{liao2025reward}; Speculative Thinking~\cite{yang2025speculative}; SCoT~\cite{wang2025efficientreasoningllmsspeculative}; SpecSearch~\cite{wang2025acceleratinglargelanguagemodel}; ValueFree~\cite{sareen2025puttingvaluerlbetter}; GG~\cite{ghasemabadi2025guidedgutefficienttesttime}; VGS~\cite{wang2025valueguidedsearchefficientchainofthought}; DO~\cite{hassid2025dontoverthinkitpreferring}; FFS~\cite{agarwal2025finishsearchefficienttesttime}; DORA~\cite{wang2025rolloutcountsoptimalresource}; SPECS~\cite{cemri2025textttspecsfastertesttimescaling}; Best-Route~\cite{ding2025bestrouteadaptivellmrouting}; LightThinker \cite{zhang2025lightthinker}; INFTYTHINK \cite{yan2025inftythink}; SCoT~\cite{xiang2025can}; RASC~\cite{wan2024reasoning}; Adaptive Reasoning~\cite{yu2025think}; AdaptiveStep~\cite{liu2025adaptivestep}; Self-Calib~\cite{huang2025efficient}; CISC~\cite{taubenfeld2025confidence}; ESC~\cite{li2024escape}; DSC~\cite{wang2024make}; PathC~\cite{zhu2024path}; RPC~\cite{zhou2025bridging}; Sleep-time Compute~\cite{lin2025sleep}; SpecReason~\cite{pan2025specreason}; TOPS~\cite{yang2025thinkingoptimalscalingtesttimecompute}; Retro-Search~\cite{lu2025retrosearchexploringuntakenpaths}; ThinkDeepFast~\cite{wang2025thinkdeepthinkfast}; AlphaOne~\cite{zhang2025alphaone}; CAR~\cite{lu2025prolongedreasoningneedcertaintybased}; Collaborative~\cite{lee2025collaborativellminferenceplanning}; STAND~\cite{song2025accelerated}; GoGI~\cite{zhuang2025accelerating}; FracCoT~\cite{liao2025fractured}; FlashThink~\cite{jiang2025flashthink}; RPC~\cite{song2025reasoning}; ThinkLess~\cite{li2025thinkless}; 
                        Plan and Budget~\cite{lin2025plan}; TrimR~\cite{lin2025trimr}; CoThink~\cite{fan2025cothinktokenefficientreasoninginstruct}; AnswerConvergence~\cite{liu2025answerconvergencesignalearly}; NOWAIT~\cite{wang2025waitdontneedwait}; Self-Affirmation~\cite{liu2025efficientreasoningsuppressionselfaffirmation}; 
                        BudgetGuidance~\cite{li2025steeringllmthinkingbudget}; Self-Guided~\cite{zhao2025exploringexploitinginherentefficiency}; ASC~\cite{azizi2025activationsteeringchainofthoughtcompression}; MUR~\cite{yan2025murmomentumuncertaintyguided}; 
                        R-Stitch~\cite{chen2025rstitchdynamictrajectorystitching}; 
                        TTPI~\cite{yang2025testtimepromptintervention}; 
                        CGRS~\cite{huang2025efficientreasoninglargereasoning}; 
                        },
                        OERleaf node]
                    ]
                ]
                [\rotatebox{0}{\parbox{3cm}{\centering Input Prompts-based \\ Efficient Reasoning}}, ERI node
                    [\parbox{3cm}{\centering Prompt-Guided Efficient Reasoning}, ERI node
                        [\parbox{15cm}{\raggedright
                        e.g. 
                        Token-Budget~\cite{han2024token}; Chain of Draft~\cite{xu2025chain}; Token Complexity~\cite{lee2025well}; Concise Chain-of-Thought (CCoT)~\cite{renze2024benefits}; MARP~\cite{chen2024unlocking}; ThoughtMani~\cite{liu2025thoughtmanipulationexternalthought}; NoThinking~\cite{ma2025reasoning}; Brevity~\cite{poddar2025brevitysoulsustainabilitycharacterizing}; PREMISE~\cite{yu2025premisescalablestrategicprompt}; ConciseHint~\cite{tang2025concisehintboostingefficientreasoning}; 
                        },
                        ERIleaf node]
                    ]
                    [\parbox{3cm}{\centering Routing by Question Attributes}, ERI node
                        [\parbox{15cm}{\raggedright
                        e.g.
                        Claude 3.7 Sonnet~\cite{anthropic_claude_sonnet}; SoT~\cite{aytes2025sketch}; Self-REF~\cite{chuang2025learningroutellmsconfidence}; Confident~\cite{chuang2025confidentseekstrongerexploring}; RouteLLM~\cite{ong2025routellmlearningroutellms}; THOUGHTTERMINATOR~\cite{pu2025thoughtterminatorbenchmarkingcalibratingmitigating}; ThinkSwitcher~\cite{liang2025thinkswitcherthinkhardthink}; SwitchCoT~\cite{zhang2025longshortcotinvestigating};  SynapseRoute~\cite{zhang2025synapserouteautorouteswitchingframework}; 
                        },
                        ERIleaf node]
                    ]
                ]
                [\rotatebox{0}{\parbox{3cm}{\centering Efficient Data and Models}}, CP node
                    [\parbox{3cm}{\centering Less Training Data}, CP node
                        [\parbox{15cm}{\raggedright
                        e.g. LIMO~\cite{ye2025limoreasoning}; s1~\cite{muennighoff2025s1simpletesttimescaling}; S$^2$R~\cite{ma2025s2rteachingllmsselfverify}; Light-R1~\cite{wen2025light}; 
                        },
                        CPleaf node]
                    ]
                    [\parbox{3cm}{\centering Pruning \& Quantization \& Distillation}, CP node
                        [\parbox{15cm}{\raggedright
                        e.g. Struggle\cite{li2025small}; Strong Verifiers~\cite{srivastava2025towards}; TinyR1-32B-Preview~\cite{sun2025tinyr1}; Mixed Distillation~\cite{chenglin2024mixed}; Counterfactual Distillation~\cite{feng2024teaching}; Feedback-Driven Distillation~\cite{zhu2024improving}; SKIntern~\cite{liao2025skintern}; Adaptive Thinking~\cite{chen2024distilling}; PRR~\cite{zhao2024probe}; CompressionReasoning~\cite{zhang2025reasoning}; TwT~\cite{xu2025twtthinkingtokenshabitual}; 
                        },
                        CPleaf node]
                    ]
                ]
                [\rotatebox{0}{\parbox{3cm}{\centering Benchmark \& Insights}}, BI node
                    [\parbox{3cm}{\centering Evaluation \& Benchmarks}, BI node
                        [\parbox{15cm}{\raggedright
                        e.g. 1B vs. 405B~\cite{liu20251bllmsurpass405b}; Sys2Bench~\cite{parashar2025inferencetimecomputationsllmreasoning}; Danger~\cite{cuadron2025dangeroverthinkingexaminingreasoningaction}; Inference-time Computation~\cite{liu2025bag}; Impact~\cite{jin2024impact}; S1-Bench~\cite{zhang2025s1}; CompressionReasoning~\cite{zhang2025reasoning}; QuantRM~\cite{liu2025quantizationhurtsreasoningempirical}; SmallRM~\cite{zhuang2025technicalstudy05breasoning}; 
                        }, 
                        BIleaf node]
                    ]
                ]
            ]
        \end{forest}
    } 
    \caption{\label{fig:Taxonomy} Taxonomy of existing literature on efficient reasoning for LLMs.}
\end{figure}
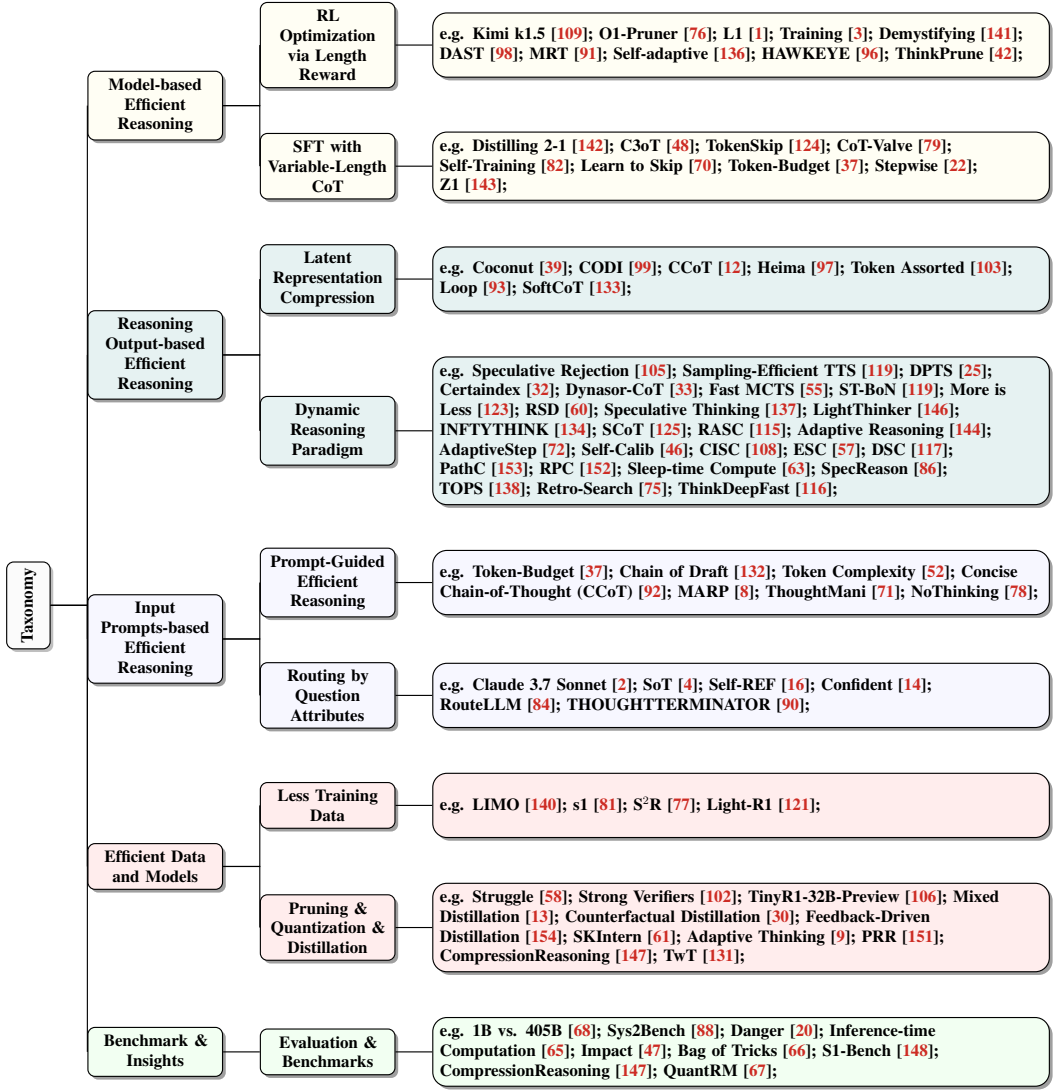

%% file: sec/2_background.tex
\section{Background: Long CoT Reasoning Models and Overthinking Phenomenon}

\subsection{Chain-of-Thought (CoT) Reasoning}

Chain-of-Thought (CoT) reasoning~\cite{wei2022chain} is a key approach that has been purposefully introduced in LLMs to enhance their reasoning capabilities. In this setting, models are typically prompted to generate a structured reasoning chain before arriving at a final answer. Techniques in this domain have been shown to improve overall accuracy~\cite{wei2022chain} since a higher-quality generation context often leads to more consistent and reliable final results. Several notable CoT variants have been developed: Self-Consistency CoT~\cite{wang2023self} replaces the standard greedy decoding approach by sampling diverse reasoning paths and selecting the most consistent answer through marginalization and aggregation. Tree-of-Thought (ToT) prompting~\cite{yao2023tree} further structures the reasoning process as a tree with backtracking, significantly improving efficiency in solving parallelizable subtasks. Graph-of-Thoughts (GoT) prompting~\cite{besta2024graph} extends this concept by structuring thoughts into a graph, allowing iterative refinement of individual reasoning steps. While many CoT variants exist, they generally involve different prompting techniques to guide the behavior of models, sometimes incorporating controller-like mechanisms to manage thought progression and usage.

\subsection{The Mechanism Behind Large Reasoning Models}

Multi-step reasoning refers to the ability of LLMs ability to generate structured reasoning steps before committing to a final answer. This capability is particularly beneficial for logic-intensive tasks such as mathematics and programming. More broadly, reasoning-capable models are often favored by human users over their non-reasoning counterparts, as evidenced by rankings in the Chatbot Arena LLM Leaderboard.\footnote{A community-driven evaluation of leading LLMs and AI chatbots: \url{https://lmarena.ai/?leaderboard}.}

Recent reasoning models, such as DeepSeek-R1~\cite{guo2025deepseek} and OpenAI o1~\cite{openai_learning_to_reason}, are known or believed to have internalized reasoning behaviors, reducing reliance on explicit test-time augmentations. These models generate detailed CoT reasoning by iteratively producing intermediate steps and refining solutions sequentially until reaching a final answer. Unlike traditional CoT approaches, which rely on prompting, these reasoning models internalize their reasoning capability through extensive training.

The OpenAI o1 model is speculated to employ a tree-based search approach, such as Monte Carlo Tree Search (MCTS)~\cite{kocsis2006bandit, coulom2006efficient}, combined with a Process Reward Model (PRM) to explore reasoning paths and determine optimal solutions through guided simulations.\footnote{There is no official confirmation regarding OpenAI o1’s training details and mechanisms. However, sources such as \url{https://www.interconnects.ai/p/openais-o1-using-search-was-a-psyop} and \url{https://www.youtube.com/watch?v=6PEJ96k1kiw} discuss these speculations in detail and are recommended for interested readers.} DeepSeek-R1, on the other hand, explicitly learns its reasoning capability through supervised fine-tuning and reinforcement learning, with a particular emphasis on rule-based rewards for math and coding tasks. These models are trained to generate reasoning steps in a predefined format before arriving at their final answers.

\subsection{The Overthinking Problem in Long CoT Reasoning Models}

\begin{figure}[t]
    \centering
    \includegraphics[width=0.99\linewidth]{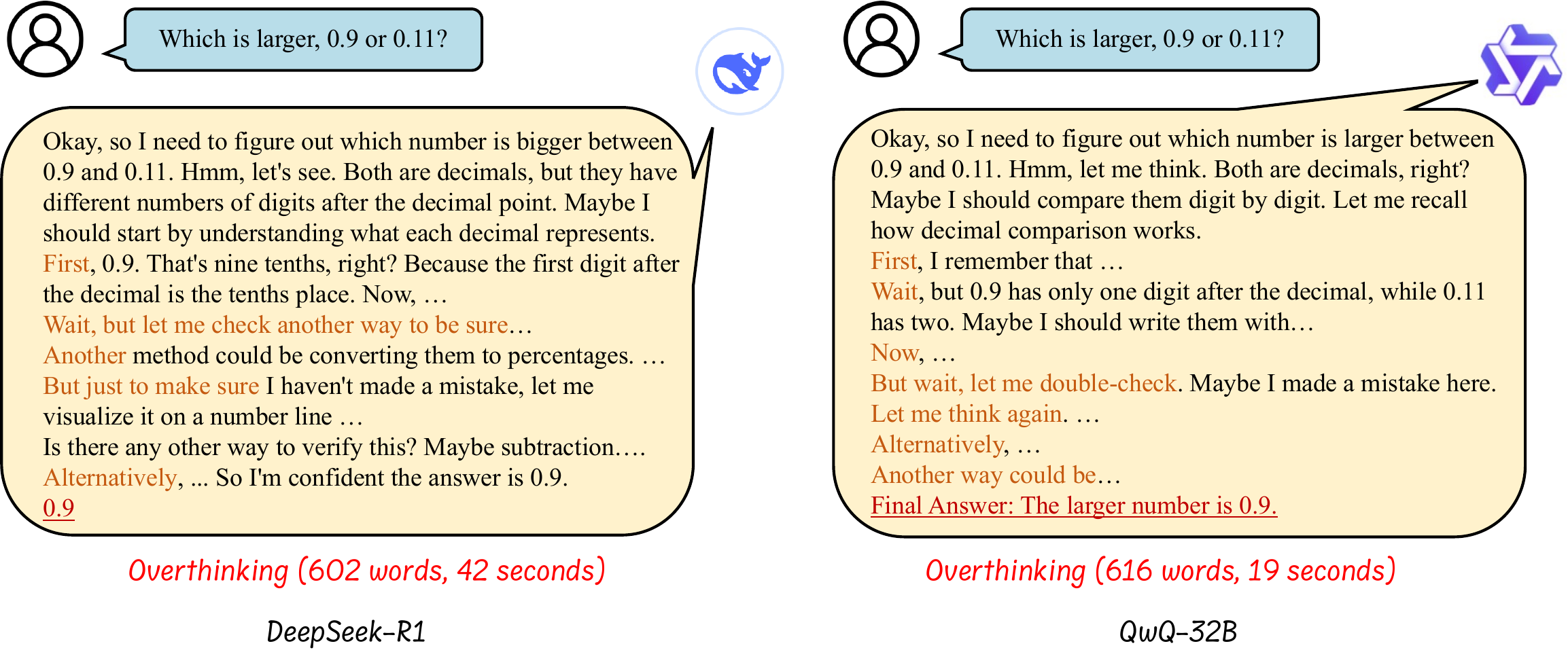}
    \caption{An example of the ``overthinking phenomenon'': when asked \textbf{\textit{``Which is larger, 0.9 or 0.11?''}}, the reasoning model takes an unnecessarily long time (e.g., 19 seconds for QwQ-32B~\cite{qwen_qwq_32b_preview} and 42 seconds for DeepSeek-R1~\cite{guo2025deepseek}) to arrive at the correct answer. This example was tested in March 2025.}
    \label{fig:overthink-generation}
\end{figure}

The ``overthinking phenomenon''~\cite{team2025kimi,chen2024not} in long CoT reasoning models refers to situations where LLMs generate excessively detailed or unnecessarily elaborate reasoning steps, ultimately reducing their problem-solving efficiency. In particular, many modern reasoning models, especially those with smaller parameter scales, tend to produce verbose reasoning or redundant intermediate steps, making them unable to provide answers within the user-defined token budget. In worse cases, excessive reasoning steps introduce errors or obscure logical clarity, leading to incorrect answers.

Figure~\ref{fig:overthink-generation} illustrates an example of overthinking. Even though the model arrives at the correct answer early in its reasoning process, it continues generating unnecessary intermediate steps, leading to inefficiencies. Given the substantial resource costs associated with LLM inference (e.g., OpenAI o1 costs \$60 per 1M generated tokens), such behavior is highly undesirable. Moreover, the problem becomes even worse if longer reasoning leads to wrong answers. In contrast, efficient reasoning models would use fewer reasoning steps to obtain correct answers while reducing inference costs.

Addressing this challenge is particularly difficult because the pretraining recipes for reasoning-capable models often explicitly encourage generating extended reasoning steps to improve accuracy. For example, DeepSeek-R1-Zero, a more or less a development prototype of DeepSeek-R1, exhibits a direct correlation between increased training duration with longer response lengths and improved benchmark performance~\cite{guo2025deepseek}. These trends are often viewed as proxies for successful reasoning training. Consequently, improving inference efficiency requires working against certain pretraining objectives, making it a non-trivial challenge. 

This paper aims to systematically summarize various approaches and methodologies toward achieving the challenging yet valuable goal of developing reasoning models with high efficiency and strong reasoning capabilities.

%% file: sec/part1_RL.tex
\section{Model-based Efficient Reasoning}

From the model perspective, these works focus on fine-tuning LLMs to improve their intrinsic ability to reason concisely and efficiently.

\subsection{RL with Length Reward Design}
\label{sec:rl}

\begin{figure}[h]
    \centering
    \includegraphics[width=\linewidth]{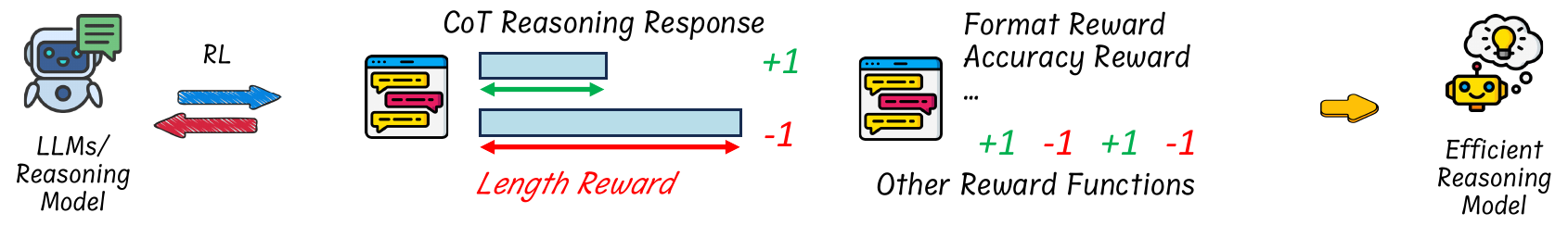}
    \caption{Illustration of the method for RL fine-tuning with length reward designs. In principle, the length reward assigns higher rewards to short, correct answers and penalizes lengthy or wrong answers to achieve efficient reasoning LLMs.}
    \label{fig:rl}
\end{figure}

Most reasoning models are trained using RL-based methods (e.g., DeepSeek-R1~\cite{guo2025deepseek}, DeepSeek-R1-Zero~\cite{guo2025deepseek}, OpenAI o1~\cite{openai_learning_to_reason}, QwQ-32B-Preview~\cite{qwen_qwq_32b_preview}) which focus on the accuracy reward and format rewards~\cite{guo2025deepseek}. To enhance reasoning-length efficiency, some studies propose integrating a length reward into the RL framework, which effectively shortens the reasoning process (as shown in Table~\ref{fig:rl}). In principle, the length reward assigns higher scores to short, correct answers while penalizing lengthy or incorrect ones, thereby optimizing the length of the reasoning path.

\begin{table}[t]
\centering
\small
\caption{Comparison of different length reward-based RL methods. $L(\cdot)$ denotes the way of calculating the prediction length. $r_0^c/r_0^w$ denotes reward (correct/wrong) for $L(\cdot)$=0. $r_L^c/r_L^w$ Reward (correct/wrong) for $L(\cdot) = L_\text{max}(\cdot)$. $r_e$ is the exceed length penalty. $y_\text{GT}$ represents the ground truth answer of input data $x$. $\pi_{\text{ref}}$ is the policy of reference model.}
\label{tab:rl-length}
\resizebox{\textwidth}{!}{
\begin{tabular}{l|c|c|c|c}
\toprule
\textbf{Method} & \textbf{RL} &  \textbf{Length Constraint Reward} & \textbf{Data} & \textbf{Model} \\
\midrule\midrule
O1-Pruner~\cite{luo2025o1} & PPO & $\mathbb{E}_{x \sim D} \left[ \mathbb{E}_{\pi_\theta, \pi_{\text{ref}}}\left[\frac{L(y_\text{ref})}{L(y_\text{pred})}\right] - 1 \right]
$ & \makecell[c]{GSM8K \\ GaoKao \\ MATH-500 } & \makecell[c]{Marco-o1-7B \\ QwQ-32B-Preview} \\
\midrule 
Demystifying~\cite{yeo2025demystifying} & PPO & $\begin{cases}
            \; r_0^c + 0.5 \times (r_L^c - r_0^c)(1 + \cos({\frac{\pi L(y_\text{pred})}{L_{\max}}}), & \text{if correct}, \\
            \; r_0^c + 0.5 \times (r_L^w - r_0^w)(1 + \cos({\frac{\pi L(y_\text{pred})}{L_{\max}}}), & \text{if wrong} \\
            \; r_e, & \text{if } L(y_\text{pred}) = L_{\max},\end{cases}$
        & \makecell[c]{MATH-500 \\ AIME-2024 \\ TheoremQA \\  MMLU-Pro-1k } & \makecell[c]{LLaMA-3.1-8B \\ Qwen2.5-7B-Math} \\
\midrule
L1~\cite{aggarwal2025l1} & GRPO & $\begin{cases}
            \; x_\text{new} = \textit{CONCAT}~(x,\textit{“Think for N tokens.”}),\\
            \; r(y, y_{GT}, L(y_{GT})) = \mathbb{I}(y_\text{pred} =  y_{GT}) - \alpha \cdot |L(y_{GT}) - L(y_\text{pred})|
        \end{cases}$ 
       & \makecell[c]{AMC \\ GPQA \\ LAST \\ MMLU \\ MATH-500 \\ AIME-2024 \\ Olympiad-Bench  } & DeepSeek-R1-Distill-Qwen-1.5B \\
\midrule
DAST~\cite{shen2025dast} & SimPO & \makecell[c]{Trained with constructed length preference data} &  \makecell[c]{MATH-500 \\ AIME-2024} & \makecell[c]{DeepSeek-R1-Distill-Qwen-7B \\ DeepSeek-R1-Distill-Qwen-32B} \\
\midrule 
Training~\cite{arora2025training} & PG & $\mathbb{E}_{x \sim D}\left[ \mathbf{1}\{y_\text{pred} = y_\text{GT}\}(1 - \alpha f(L(y_\text{pred}))) \right]$ & \makecell[c]{GSM8K \\ MATH-500 \\ AIME-2024  } & \makecell[c]{DeepSeek-R1-Distill-Qwen-1.5B \\ DeepSeek-R1-Distill-Qwen-7B} \\
\bottomrule 
\end{tabular}
}
\end{table}

\begin{table}[t]
\centering
\small
\caption{Comparison of different policy optimization methods in CoT length controls. $\hat{R}_t$ represents the reward model. $\pi_{\text{ref}}$ is the policy of reference model. $\gamma$ is a target reward margin term for SimPO. $\lambda$ is a clipping-related hyper-parameter. The $y_w$ is for winning responses, and $y_l$ is for losing responses, where some with $G$ on superscript denote the outputs of different sampled groups.}
\label{tab:rl-comparison}
\scalebox{0.9}{
\begin{tabular}{l|c}
\toprule
\textbf{Method} & \textbf{Optimization Objective} \\
\midrule\midrule
Policy Gradient (PG) & $\mathbb{E}_{\pi_\theta}\left[ \nabla_\theta \log \pi_\theta(y_t|x_t) \hat{R}_t \right]$ \\
\midrule
PPO~\cite{schulman2017proximal} & $\mathbb{E}\left[ \min \Big( 
\frac{\pi_\theta(y_t  \mid x_t)}{\pi_{\theta_{\text{ref}}}(y_t  \mid x_t)}\hat{R}_t ,\ 
\text{clip}\left( 
\frac{\pi_\theta(y_t \mid x_t)}{\pi_{\theta_{\text{ref}}}(y_t \mid x_t)},\ 
1 - \epsilon,\ 
1 + \epsilon 
\right)\hat{R}_t  
\Big)
\right]$ \\
\midrule
SimPO~\cite{meng2024simpo} & $\mathbb{E}\left[ \log \sigma \left(
\frac{\beta}{|y_t^w|} \log \pi_\theta(y_t^w \mid x_t)
- \frac{\beta}{|y_t^l|} \log \pi_\theta(y_t^l \mid x_t)
- \gamma
\right) \right]$\\
\midrule
GRPO~\cite{shao2024deepseekmath} & $\mathbb{E}\left[ \min \Big( 
\frac{\pi_\theta(y^G_t  \mid x_t)}{\pi_{\theta_{\text{ref}}}(y^G_t  \mid x_t)}\hat{R}^G_t ,\ 
\text{clip}\left( 
\frac{\pi_\theta(y^G_t \mid x_t)}{\pi_{\theta_{\text{r}}}(y^G_t \mid x_t)},\ 
1 - \epsilon,\ 
1 + \epsilon 
\right)\hat{R}^G_t  
\Big) - \lambda\mathbb{D}_{KL}[\pi_\theta||\pi_{\text{ref}}] \right]$ \\
\bottomrule
\end{tabular}
}
\end{table}

\insightbox{The key question is: How to formulate the length reward in RL?}

Existing works leverage traditional RL optimization techniques combined with \textbf{\textit{explicit length-based reward}} to control the length of CoT reasoning. Some detailed length rewards are shown in Table~\ref{tab:rl-length}. 
The work~\cite{arora2025training} proposes utilizing length-based rewards conditioned on correctness, where shorter correct answers receive higher rewards. They then apply traditional policy gradient methods guided by this reward scheme to encourage LLMs to produce concise reasoning steps.
Expanding from the policy gradient, the following discussed work is primarily built upon proximal policy optimization (PPO)~\cite{schulman2017proximal} with CoT length penalty.
Demystifying~\cite{yeo2025demystifying} presents empirical findings from RL experiments examining how reasoning capability is influenced by length. They demonstrate that RL does not consistently or reliably increase the length and complexity of CoT reasoning, emphasizing the necessity of controlling CoT length growth to ensure stable performance. To mitigate these issues, they proposed a Cosine Reward based on a Dirichlet function of concise reward formula~\cite{loshchilov2016sgdr} and the proposed ``exceed length penalty'' scores.
Due to the performance impact of CoT length, Kimi k1.5~\cite{team2025kimi} incorporates a length penalty into its policy optimization (a variant of online policy mirror decent~\cite{tomar2020mirror}) to improve long CoT activations and facilitate effective model merging. Besides optimizing with length penalty reward, L1~\cite{aggarwal2025l1} modify the training data with the designated length constraint instruction (i.e., Think for $N$ tokens) before launching the policy optimization with pre-trained reasoning LLMs. 
O1-Pruner~\cite{luo2025o1} introduces the Length-Harmonizing Reward, combined with a PPO-style loss, to optimize reasoning LLMs by effectively shortening the CoT length. Specifically, the Length-Harmonizing Reward is computed based on the ratio of CoT lengths between the reference model output and the predicted results. Additionally, this reward incorporates accuracy-based constraints comparing predictions to the reference model outputs, ensuring that shortening the reasoning process does not degrade task performance.
Without relying on a reference model, DAST~\cite{shen2025dast} employs SimPO~\cite{meng2024simpo} to fine-tune reasoning LLMs using a constructed length-preference dataset. This dataset is generated based on a self-defined token-length budget measurement $L_\text{budget}$, defined as a linear combination of the average token length of correct responses and the maximum allowed generation length.

These RL-based methods enable the mitigation of overthinking in reasoning-capable LLMs, where overthinking refers to unnecessarily extended reasoning processes, leading to longer inference times and exceeding computational budgets. By achieving nearly lossless alignment with the original reasoning capabilities of LLMs, these budget-efficient RL strategies democratize the deployment of reasoning LLMs in resource-constrained scenarios.

%% file: sec/part2_data.tex
\subsection{SFT with Variable-Length CoT Data}
\label{sec:longshortdata}

\begin{figure}[h]
    \centering
    \includegraphics[width=\linewidth]{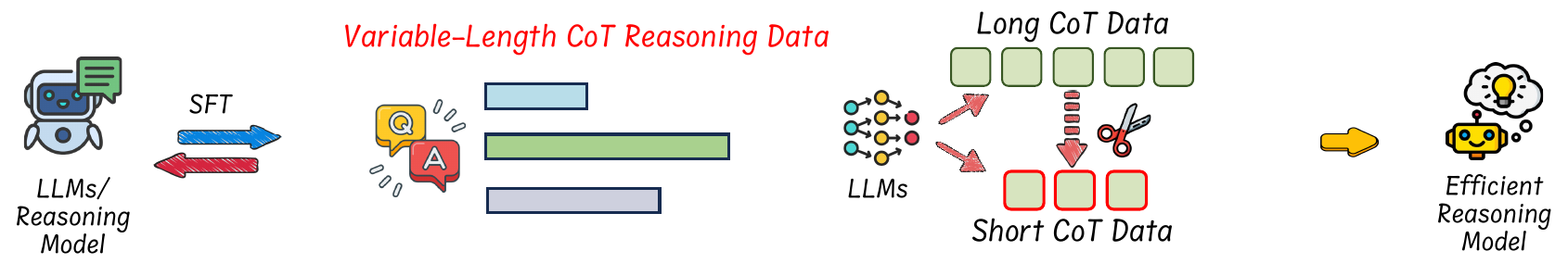}
    \caption{Illustration of methods for utilizing SFT with variable-length CoT reasoning datasets.}
    \label{fig:sft}
\end{figure}

\textbf{\textit{Fine-tuning LLMs with variable-length CoT data}} is an effective way to improve the efficiency of reasoning. As shown in Figure~\ref{fig:sft}, this series of works typically involves: (1) Constructing variable-length CoT reasoning datasets via various methods, and (2) Applying SFT with collected data on reasoning models to enable LLMs to learn compact reasoning chains that encapsulate effective knowledge. Note that this method is not limited to RL-trained reasoning models; it can also directly enhance reasoning models by injecting efficient reasoning capabilities, similar to those used in distilled reasoning models.(e.g., DeepSeek-R1-Distill-Qwen~\cite{guo2025deepseek}).

\insightbox{The key question is: How to collect variable-length CoT reasoning data, especially for short CoT data?}

\begin{table}[ht]
\centering
\small
\caption{Comparison of various approaches that utilize SFT with variable-length CoT reasoning datasets.}
\label{tab:variable-length}
\resizebox{\textwidth}{!}{
\begin{tabular}{l|cccccccccc}
\toprule
\textbf{Method} & \textbf{Source Data} & \textbf{Reasoning Pruning} & \textbf{SFT} & \textbf{LLMs} \\
\midrule\midrule
Self-Training~\cite{munkhbat2025self} & \makecell[c]{GSM8K \\ MATH} & \makecell[c]{Sampling $N$ \\ reasoning then select \\ the shortest one}  & Standard & \makecell[c]{Llama-3.2-\{1B,3B\} \\ Llama-3.1-8B} \\ 
\midrule
TokenSkip~\cite{xia2025tokenskip} & \makecell[c]{GSM8K \\ MATH} & \makecell[c]{Skip tokens according to \\ semantic importance} & Standard & \makecell[c]{LLaMA-3.1-8B-Instruct \\ Qwen2.5-Instruct} \\
\midrule
C3oT~\cite{kang2024c3ot} & \makecell[c]{GSM8K \\ MathQA \\ ECQA \\ StrategyQA} & \makecell[c]{GPT-4 as compressor \\ to make concise \\ reasoning} & Standard & Llama-2-chat-\{7B,13B\} \\
\midrule
Distilling2-1~\cite{yu2024distilling} & OASST2 & Removing reasoning & Standard & Llama-2-70B-chat \\
\midrule
Token-Budget~\cite{han2024token} & \makecell[c]{GSM8K \\ GSM8K-Z \\ MathBench} & \makecell[c]{Persuing an optimal \\ token budget for LLMs \\ to complete the reasoning} & Standard & Llama-3.1-8B-Instruct \\
\midrule
CoT-Valve~\cite{ma2025cot} & \makecell[c]{GSM8K \\ PRM800k} & \makecell[c]{Merging parameters \\ of non-reasoning and \\ long reasoning LLMs} & Progressive & \makecell[c]{QwQ-32B-Preview \\ DeepSeek-R1-Distill-Llama-8B \\ LLaMA-3.1-8B \\ LLaMA-3.2-1B \\ Qwen32B-Instruct} \\
\midrule
LearnSkip~\cite{liu2024can} & \makecell[c]{Analog of Algebra \\ Multi-digit Addition \\ Directional Reasoning} & \makecell[c]{Stage 1: Manually skipping \\ Stage 2: Prompting LLMs \\ for shorter reasoning} & \makecell[c]{Standard \& \\ Progressive} & \makecell[c]{Llama-2-7B \\ Phi-3-mini~(3.8B)} \\
\bottomrule
\end{tabular}
}
\end{table}

\subsubsection{Constructing Variable-Length CoT Reasoning Datasets}

Variable-length CoT reasoning datasets refer to datasets of long/short reasoning steps that could guide LLMs to achieve correct answers.
Existing works typically gather long CoT data by prompting pre-trained reasoning models with questions. Based on the long CoT data, the key challenge is: \textit{How to collect short CoT data?} Overall, variable-length CoT reasoning datasets can be created via either post-reasoning or during-reasoning. We list some detailed approaches in Table~\ref{tab:variable-length}.

\paragraph{\textbf{Post-reasoning CoT Compression.}}
This approach collects short CoT data by reducing redundant reasoning steps after full-length reasoning, either by heuristic criterion or LLMs, as proposed in \cite{yu2024distilling},~\cite{kang2024c3ot}, and~\cite{xia2025tokenskip}.  
Specifically, \cite{yu2024distilling} uses reasoning-capable LLMs to generate the reasoning and answers. After generating full-length CoT data, they discard the reasoning process, only using the questions and answers to distill system-1 LLMs.
Another work C3oT improves the reasoning efficiency by compressing the reasoning process~\cite{kang2024c3ot}. The long CoT reasoning steps were generated by explicitly prompting LLMs. Then, it employs GPT-4 as a compressor to reduce the length of the reasoning process while ensuring the compressed reasoning retains all key information and removes redundant words. In addition, TokenSkip reduce the reasoning steps driven by interpretation~\cite{xia2025tokenskip}.
It estimates the semantic importance of each reasoning part to the final answer and reduces the reasoning tokens.
The important parts preserve the key reasoning steps that could improve the accuracy of the final answer.
The advantage of post-reasoning CoT compression is that it can achieve a higher reduction rate of the reasoning steps, which advances more efficient reasoning.

\paragraph{\textbf{Obtaining Compressed CoT Data during Reasoning.}}
This approach collects short CoT data by prompting LLMs to generate short reasoning steps during inference and reasoning, as proposed in \cite{liu2024can}, \cite{munkhbat2025self},~\cite{han2024token}, and~\cite{ma2025cot}.
Specifically, \cite{liu2024can} proposes a human-like step-skipping method for generating shorter reasoning steps. In the first stage, based on the original training datasets, they manually create solutions by skipping steps, either guided by human expertise or by randomly merging or removing steps. Further, these concise data are labeled with prompts such as “Solve it in $n$ steps.”. After SFT, the model is able to generate shorter reasoning paths. In the second stage, they prompt this model to solve problems by intrinsically skipping or compressing steps during reasoning.
The generated concise reasoning steps with questions and answers are collected as datasets, which are then used in SFT to make LLMs solve problems with fewer steps.
Moreover, Token-Budget~\cite{han2024token} has an important insight: an optimal token budget helps LLMs actively follow the token constraint to complete the reasoning process.
Motivated by this insight, it proposes a binary search-based method to achieve the optimal token budgets, and follow these budgets to generate short reasoning steps. 
In addition, \cite{munkhbat2025self} proposes a sampling-based method to improve reasoning efficiency.
Specifically, it examines the distribution of reasoning lengths and finds that shorter solutions appear more frequently than the typical reasoning length.
Driven by this finding, it proposes a Best-of-N~(BoN) Sampling at test time, which generates $N$ paths of reasoning and selects the shortest one.
These short reasoning paths are collected as the dataset.
Finally, CoT-Valve~\cite{ma2025cot} controls the reasoning length by mix-up the parameters of long reasoning and non-reasoning LLMs for generating variable-length reasoning steps.
The advantage of CoT compression during reasoning is that the naturally generated reasoning steps align with the intrinsic knowledge of LLMs, which advances more effective learning of LLMs.

\subsubsection{Fine-Tuning Approaches}

After collecting variable-length CoT data, existing works fine-tune LLMs to achieve efficient reasoning in several ways, which include standard fine-tuning (e.g., parameter-efficient fine-tuning such as LoRA~\cite{hu2022lora} or full fine-tuning) and progressive fine-tuning.

\paragraph{\textbf{Standard Fine-tuning.}}
Most of the work adopts standard methods to fine-tune LLMs~\cite{liu2024can, munkhbat2025self, yu2024distilling, kang2024c3ot, xia2025tokenskip, han2024token}. 
Specifically, these approaches adopt LoRA~\cite{hu2022lora} or full fine-tuning~\cite{kang2024c3ot} to minimize the perplexity loss function or DPO loss function~\cite{han2024token} on the reasoning-efficient datasets.
The LoRA enables LLMs to adapt to short reasoning steps with less than $1\%$ of the parameters tuned.
In addition, \cite{liu2024can} observed the growing reasoning efficiency can generalize to out-of-domains beyond the collected datasets.

\paragraph{\textbf{Progressive Fine-tuning.}}
Progressive fine-tuning aims to smoothly reduce the reasoning steps during fine-tuning~\cite{ma2025cot, liu2024can}.
One way is to progressively reduce the reasoning steps of data during fine-tuning LLMs, as employed in \cite{liu2024can}.
Another effective way is to progressively adjust the generation of reasoning steps, as proposed by CoT-Valve~\cite{ma2025cot}. 
Specifically, it first learns LoRA adaptor $\Delta \theta_{\text{N}}$ and $\theta_{\text{L}}$, where LLMs with $\Delta \theta_{\text{N}}$ have no reasoning steps, and that with $\Delta \theta_{\text{L}}$ have long reasoning.
Then, it mix-up $\Delta \theta_{\text{N}}$ and $\Delta \theta_{\text{long}}$ by  
$\alpha \Delta \theta_{\text{N}} + (1-\alpha) \Delta \theta_{\text{L}}$ to generate a dataset reasoning with variable length.
Here $0 < \alpha < 1$ controls the parameter to shift from $\Delta \theta_{\text{N}}$ to $\Delta \theta_{\text{L}}$, controlling the reasoning length generated by LLMs.
Finally, it fine-tunes LLMs on the generated data while progressively reducing $\alpha$ from 1 to 0.
In this way, reasoning efficiency is progressively improved during fine-tuning.

\paragraph{Model Merging.} Beyond fine-tuning reasoning models using the above CoT data, several works have explored model merging to improve reasoning efficiency in LLMs. Kimi k1.5~\cite{team2025kimi} investigates merging strategies~\cite{yang2024modelmergingllmsmllms} that transform models producing lengthy CoT traces into ones that generate shorter, more concise reasoning outputs. Unlocking~\cite{wu2025unlockingefficientlongtoshortllm} conducts an empirical study on model merging for efficient reasoning with various methods, including task-vector-based, SVD-based, and activation-informed merging. 

%% file: sec/part3_latent.tex
\section{Reasoning Output-based Efficient Reasoning}
From the perspective of reasoning steps in the output, these works focus on modifying the output paradigm to enhance the ability of LLMs to reason concisely and efficiently.

\subsection{Compressing Reasoning Steps into Fewer Latent Representation}
\label{sec:latent}

Although standard CoT methods improve LLM performance by explicitly writing reasoning steps, recent work~\cite{deng2024explicit} has shown that simply adding intermediate ``thinking'' tokens, or even meaningless filler (e.g., ``......'')~\cite{pfau2024let}, can also increase performance. \cite{geiping2025scaling} scales up deeper reasoning through recurrent expansions in the hidden space rather than verbose text. These findings highlight that the benefit often lies in more hidden computation rather than purely textual decompositions. Building on the insight that latent reasoning can allow LLMs to reason more efficiently and flexibly, \textit{with fewer (or no) explicit textual intermediate steps, several new methods focus on compressing or replacing explicit CoT with more compact latent representations.}

\insightbox{The key question is: How to compress reasoning steps into latent space?}

\begin{figure}[h]
    \centering
    \includegraphics[width=\linewidth]{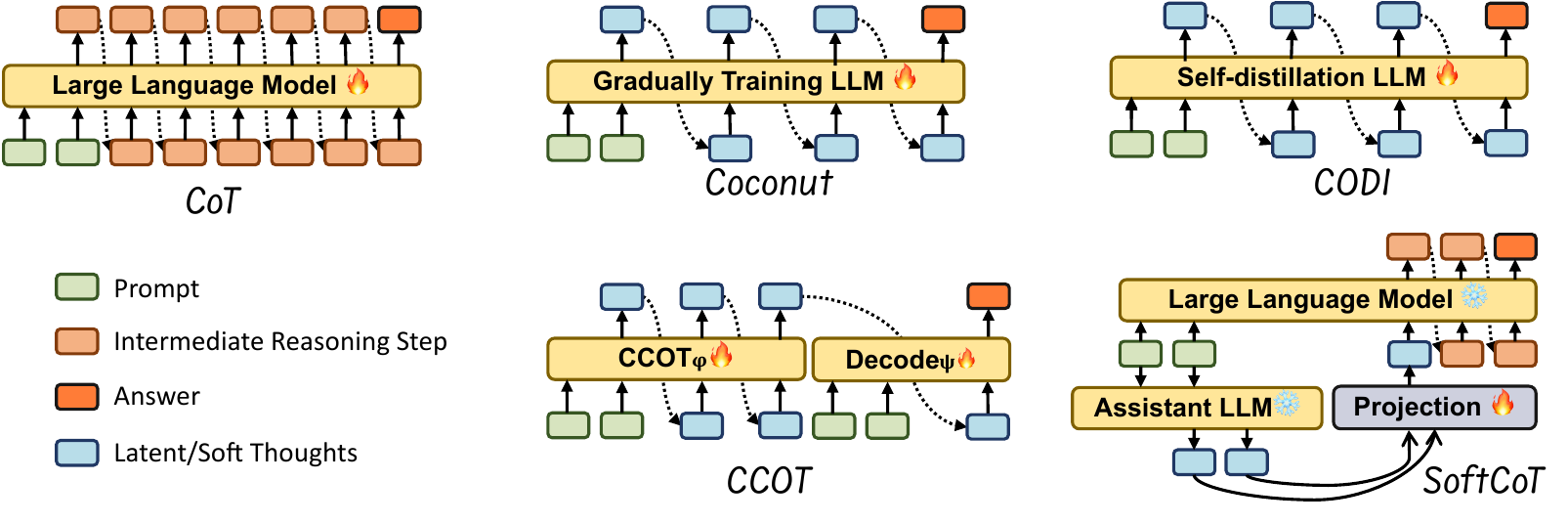}
    \caption{
    Comparison of methods of compressing reasoning steps into fewer latent representations.}
    \label{fig:latent}
\end{figure}

In general, these methods can be categorized into two types: training LLMs to inference using latent representations or using an auxiliary model. A visualized comparison of some of these approaches is presented in Figure~\ref{fig:latent}.

\paragraph{Training LLMs to Leverage Latent Representations.}
Among the first explorations, Coconut (Chain of Continuous Thought)~\cite{hao2024training} treats the final-layer hidden states of an LLM as ``continuous thought'' to replace traditional discrete tokens. It then reuses these hidden states as the next input embeddings. Trained step by step, Coconut gradually adds these latent CoT tokens. The results suggest that compressing tokens into latent representations improves both accuracy and efficiency by reducing the number of intermediate ``thinking'' tokens.
CODI~\cite{shen2025codi} leverages a different training process compared to Coconut, which learns the continuous latent CoT via \textit{self-distillation}. In CODI, the model serves both teacher and student, jointly learning explicit and implicit CoT while aligning hidden activations on the token, generating the final answer. This self-distillation process enables LLMs to perform reasoning internally without generating explicit CoT tokens.
Similarly, CCOT~\cite{cheng2024compressed} condenses long CoT reasoning into short \textit{contentful and continuous contemplation tokens}. First, it precomputes the full CoT for a query and selects the most important hidden states as a gold standard for compression. The CCOT module (a LoRA) is trained to predict these key tokens. Then, the DECODE module (another LoRA) is trained on the query plus compressed tokens. During inference, CCOT generates compressed tokens, which DECODE uses to produce concise reasoning steps. Another type of work, summarization-based dynamic reasoning, as mentioned in Section~\ref{sec:dynamic} explores compressing and summarizing reasoning steps in discrete space during inference, which is similar to the introduction of ``contemplation token''.

Another work, Heima~\cite{shen2025efficient}, inspired by Coconut~\cite{hao2024training}, brings latent reasoning into Multimodal Large Langue Models (MLLMs). Instead of always using full, lengthy reasoning explanations, Heima replaces each stage of detailed reasoning with a single ``thinking token''. With this change, the training data is updated. Instead of long textual explanations, each reasoning stage is just one of these thinking tokens. Then, they continue fine-tuning the model to achieve efficient reasoning.
Token Assorted~\cite{su2025token} adopts a hybrid approach. During training, part of the CoT is replaced by discrete latent tokens learned via a VQ-VAE~\cite{van2017neural}, and then the LLM is trained with a partial and high-level abstract of the reasoning steps. The authors show that mixing text tokens with latent tokens can facilitate training and inference by representing some reasoning steps in a compact latent form. 
Other than explicitly compressing the discrete tokens into latent space, \cite{saunshi2025reasoning} demonstrates that looping a $k$-layer transformer $L$ times can emulate the performance of a $kL$-layer model. This looping mechanism effectively increases the depth of the model depth without adding parameters, enabling iterative reasoning processes within the latent space. The study reveals that \textit{looped models implicitly generate latent thoughts}, allowing them to simulate multiple steps of CoT reasoning through successive loops.

\paragraph{Training Auxiliary Modules while Keeping LLMs Frozen.} 
While most methods for continuous-space reasoning fine-tune the pre-trained LLM, SoftCoT~\cite{xu2025softcot} keeps the underlying LLM frozen. A lightweight auxiliary model generates instance-specific soft thought tokens projected into the embedding space of the frozen LLM. Experiments show that SoftCoT consistently boosts performance, demonstrating the viability of augmenting LLMs with external latent reasoning tokens.

These methods hint at a broader move toward latent reasoning, where critical thinking occurs in compressed, non-textual forms. Such approaches can unlock improved speed, adaptive inference, parallel backtracking, and new ways to interpret or partially reveal the model reasoning. As LLMs grow larger and tasks become more complex, balancing thorough reasoning with computational efficiency is greatly beneficial from these flexible and compact latent CoT paradigms.

%% file: sec/part4_dynamic.tex
\subsection{Dynamic Reasoning Paradigm during Inference}
\label{sec:dynamic}

Existing works focus on \textit{modifying the reasoning paradigm} for more efficient inference. The key during inference is choosing the proper criterion to guide the reasoning strategy. Current training-free approaches explore dynamic reasoning using various criteria, such as reward-guided, confidence-based, and consistency-based selective reasoning. Additionally, a summarization-based dynamic reasoning method intrinsically integrates the output summarization paradigm of LLMs during training.

\insightbox{The key question is: Which criterion to guide the inference? What is the appropriate efficient inference paradigm?}

\begin{table}[htbp]
\centering
\scriptsize
\caption{Comparison of different methods of dynamic reasoning paradigm of test time compute during inference.}
\scalebox{0.99}{
\begin{tabular}{p{1.5cm} p{2cm} p{0.8cm} p{3cm} p{4cm}}
\toprule
\textbf{Category} & \textbf{Method} & \textbf{Training-free?} & \textbf{Baseline and Its Drawbacks} & \textbf{Method Description} \\ \midrule
\multirow{2}{*}{\parbox{1.7cm}{Reward-guided Efficient Reasoning}} 
  & Speculative Rejection~\cite{sun2024fast} & Yes & \textbf{Best-of-N (BoN) Decoding}: underutilizes GPU memory and computational resources during the early stages, leading to lower final rewards. & Starts BoN with a large initial batch size and rejects unpromising sequences periodically, efficiently achieving higher rewards. \\ 
  \cmidrule(l){2-5}
  & Reward-Guided Speculative Decoding (RSD) \cite{liao2025reward} & Yes & \textbf{Speculative Decoding}: strictly enforces unbiasedness, discarding useful intermediate outputs and leading to computational inefficiency. & A speculative decoding method that leverages a reward model (PRM) to selectively accept high-quality outputs from a lightweight draft model, reducing computation while preserving accuracy. \\
  \midrule
  \multirow{4}{*}{\parbox{1.7cm}{Confidence/ Certainty-based Adaptive Reasoning}} & Dynamic Parallel Tree Search \cite{ding2025dynamic} & Yes & \textbf{Tree-of-Thoughts:} difficult to parallelize due to frequent switching of reasoning focus, and inefficient because of redundant exploration of suboptimal solutions & Dynamically parallelizes node expansion through adaptive batching and implements a search-and-transition mechanism (including \textit{Early Stop} and \textit{Deep Seek}) to prune unpromising paths early. \\
  \cmidrule(l){2-5}
  & Dynasor (Certaindex-based Scheduling) \cite{fu2024efficiently} & Yes & \textbf{Serving systems with uniform resource allocation}: allocate compute uniformly, leading to inefficient resource usage and unmet latency targets & Dynamically allocates compute for reasoning queries based on \textit{Certaindex}, a statistical measure of reasoning progress, to maximize accuracy within resource constraints. \\ 
  \cmidrule(l){2-5}
  & FastMCTS \cite{li2025fastmcts} & Yes & \textbf{Rejection Sampling}: inefficient, discards intermediate steps, and fails to balance problem difficulty & An MCTS-inspired sampling strategy that efficiently generates high-quality multi-step reasoning data, providing step-level evaluation signals and balanced sampling across problem difficulties. \\ 
  \cmidrule(l){2-5}
  & Length-filtered Vote \cite{wu2025more} & Yes & \textbf{Majority Voting}: ignores reasoning quality, includes suboptimal CoT lengths, and suffers from noisy predictions & A voting strategy that selects answers with the optimal CoT length by filtering out excessively short or long reasoning paths. \\
  \midrule
  Consistency-based Selective Reasoning & Self-Truncation Best-of-N (ST-BoN) \cite{wang2025sampling} & Yes & \textbf{Best-of-N Sampling}: fully generates all samples and relies on costly reward models & Estimates the most promising sample early via internal embedding consistency, truncating inferior samples prematurely. \\
  \midrule
  \multirow{2}{*}{\parbox{1.7cm}{Summarization-based Dynamic Reasoning}} & LightThinker \cite{zhang2025lightthinker} & No & \textbf{Chain-of-Thought (CoT)}: high memory and computational overhead due to the generation of an excessive number of tokens & Trains LLMs to learn when and how to compress intermediate thoughts, condensing long reasoning chains into gist tokens, and uses a sparse-patterned attention mask during inference to enhance computational efficiency. \\
    \cmidrule(l){2-5}
  & InftyThink \cite{yan2025inftythink} & No & \textbf{Monolithic Reasoning}: reasoning output is verbose, and can quickly exceed the context window limit of the LLM, resulting in poor performance & An iterative reasoning paradigm that interleaves reasoning steps with intermediate summarization, enabling unbounded reasoning depth without architectural modifications. \\
\bottomrule
\end{tabular}
}
\end{table}

\subsubsection{Dynamic Reasoning via Explicit Criteria}

Train-time scaling with RL~\cite{guo2025deepseek} can significantly enhance the reasoning ability of LLMs. However, it requires substantial computational resources to scale up the model training, making it prohibitively expensive~\cite{guo2025deepseek}.
As an alternative, researchers have explored test-time reasoning, also known as test-time scaling~\cite{snell2024scaling}. Instead of relying on training to learn CoT reasoning steps, test-time scaling leverages various inference strategies that allow models to ``think longer and broader'' on complex problems. This approach consistently improves performance on challenging math and code problems that require reasoning by increasing the computational resources allocated during inference~\cite{snell2024scaling, beeching2024scalingtesttimecompute}.

Test-time scaling utilizes various inference strategies to generate longer and higher-quality CoT responses. There are several ways to scale up the inference. (1) Best-of-N sampling~\cite{sun2024fast, wang2025sampling} involves generating multiple responses for a given prompt, expanding the search space to identify better solutions. After generation, the best response is selected using either majority voting, where the most frequently occurring response is chosen; or by a reward model, which evaluates response quality based on pre-defined criteria. This method has been shown to significantly enhance the reasoning capabilities of LLMs~\cite{beeching2024scalingtesttimecompute}. (2) Beam-based searching~\cite{ding2025dynamic, fu2024efficiently,beeching2024scalingtesttimecompute}, which differs from Best-of-N by structuring generation into multiple steps. Instead of generating an entire response in one pass, beam search selects the most promising intermediate outputs with process reward model~\cite{uesato2022solving} at each step, while discarding less the optimal ones. This enables a more fine-grained optimization of both response generation and evaluation. (3) Monte Carlo Tree Search (MCTS)~\cite{li2025fastmcts}, where multiple solution paths are explored in parallel. MCTS generates partial responses along different branches of a solution tree, evaluates them, and back-propagates reward values to earlier nodes. The model then selects the branch with the highest cumulative reward, ensuring a more refined selection process compared to traditional beam search.

Although test-time scaling can significantly reduce train-time scaling up overhead~\cite{beeching2024scalingtesttimecompute}, the large number of generated responses still makes inference computationally expensive. To address this, recent works have been exploring methods to optimize test-time scaling.

\begin{figure}[t]
    \centering
    \includegraphics[width=\linewidth]{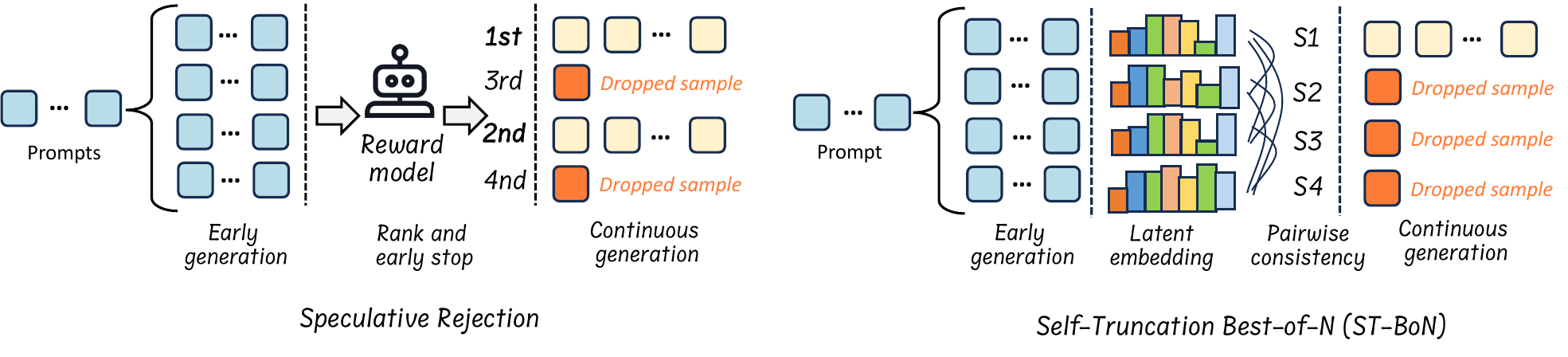}
    \caption{Examples of efficient Best-of-N sampling methods. \textit{(Left) Speculative Rejection}~\cite{sun2024fast} uses a reward model to estimate partial generation quality. It then early stops the sampled sequence with lower scores. \textit{(Right) ST-BoN}~\cite{wang2025sampling} evaluates the latent embedding of the early generation. The latent embedding of each thinking path will be used to calculate pairwise consistency between other tokens. The sequence with the highest consistency is more likely to arrive at the correct answer.}
    \label{fig:bonsample}
\end{figure}

\paragraph{Reward-guided Efficient Reasoning.}
Speculative Rejection~\cite{sun2024fast} is an efficient inference-time reasoning algorithm that optimizes Best-of-N (BoN) decoding by dynamically reducing computational overhead (as shown in Figure~\ref{fig:bonsample}, left). It generates multiple responses until memory limits are nearly reached, then \textit{discards low-quality outputs based on evaluation by a reward model}. This adaptive filtering substantially reduces inference costs compared to vanilla BoN.
On the other hand, Reward-Guided Speculative Decoding (RSD) \cite{liao2025reward} enhances the efficiency of speculative decoding specifically for multi-step reasoning tasks. Unlike traditional speculative decoding methods, which strictly require exact token matching between the draft model and target model, \textit{RSD leverages a Process Reward Model (PRM) to dynamically evaluate intermediate outputs} from the smaller, more efficient draft model. Outputs with high reward scores are directly accepted, while those with lower scores are further refined by a larger, more capable target model.

\paragraph{Confidence/Certainty-based Adaptive Reasoning.}
Dynamic Parallel Tree Search (DPTS)~\cite{ding2025dynamic} optimizes tree-based reasoning in LLMs by addressing two main inefficiencies by introducing: (1) \textit{Parallelism Streamline} optimizes memory and compute by storing only incremental KV cache updates and dynamically adjusting the number of extended nodes based on available GPU memory, (2) \textit{Search and Transition Mechanism} balances exploration and exploitation using confidence-based criteria. Overall, during inference, the system cuts off uncertain paths to save time.
FastMCTS~\cite{li2025fastmcts} is another confidence-based method that aims to optimize multi-step reasoning data synthesis. Traditional rejection sampling generates multiple candidate responses independently, selecting only the correct ones, but it is often inefficient and struggles with imbalanced sampling. Inspired by MCTS, FastMCTS prioritizes high-confidence traces for deep reasoning. Additionally, it adjusts tree expansion based on problem complexity, improving both efficiency and reasoning diversity.
Another line of research leverages certainty or uncertainty measures to guide adaptive reasoning. Certaindex~\cite{fu2024efficiently}, a certainty metric, quantifies the confidence of LLMs throughout reasoning using semantic entropy, reward model scores, or a combination of both. A higher certaindex indicates that further reasoning steps are unlikely to change the final answer, allowing early termination to free resources for more challenging queries. Dynasor, an inference system built on this principle, optimizes compute scheduling by dynamically tracking reasoning progress instead of allocating resources uniformly. Dynasor-CoT~\cite{fu2025reasoning} uses a probing scheme to exit early if the model is confident enough.
Length-filtered Vote~\cite{wu2025more} is another work that leverages uncertainty to improve CoT reasoning. The study finds that longer reasoning chains do not always improve accuracy; instead, performance initially improves but eventually declines due to error accumulation. The authors provide a mathematical analysis proving the existence of an optimal CoT length, determined by model capability and task difficulty. To exploit this, they propose Length-filtered Voting, a length-aware majority voting method that groups answers by CoT length and selects the most reliable group based on prediction uncertainty.
Self-Calib~\cite{huang2025efficient} introduces confidence-driven adaptive scaling strategies at test time to efficiently address queries with differing complexity levels, including Early-Stopping mechanisms for Best-of-N sampling and confidence-calibrated Self-Consistency approaches.
CISC~\cite{taubenfeld2025confidence} implements a weighted majority voting scheme utilizing model-derived confidence scores. By emphasizing reasoning paths with higher confidence, it efficiently determines the correct response with substantially fewer samples. 

\paragraph{Consistency-based Selective Reasoning.} Self-Truncation Best-of-N (ST-BoN)~\cite{wang2025sampling} enhances BoN sampling efficiency by introducing early termination (as shown in Figure~\ref{fig:bonsample}, right), similar to Speculative Rejection~\cite{sun2024fast}. However, unlike Speculative Rejection using reward models, ST-BoN leverages consistency as the metric to measure the importance. Specifically, it leverages the consistency of latent embeddings to evaluate response quality. The core insight is that ``the closer a sample is to others, the more likely its path will lead to the correct answer''. Then, ST-BoN selects the most consistent Chain-of-Embedding (CoE) to others and regards it as the optimal sample.

\subsubsection{Summarization-based Dynamic Reasoning}

Some existing methods choose to optimize reasoning efficiency by training LLMs to \textit{summarize intermediate thinking steps}.
LightThinker \cite{zhang2025lightthinker} proposes to train LLMs to learn when and how to compress intermediate reasoning steps. Instead of storing long thought chains, LightThinker compresses verbose reasoning into compact ``gist tokens'' to reduce memory and computational costs. Implementing this summarization paradigm requires a sparse-patterned attention mask, ensuring the model focuses only on essential compressed representations. 
InftyThink \cite{yan2025inftythink} introduces an iterative reasoning method that enables essentially infinite reasoning chains while maintaining strong accuracy without surpassing the context window limit. It achieves this by iteratively generating a thought, summarizing it, and discarding previous thoughts and summaries, retaining only the most recent summary. Additionally, InftyThink provides a technique for converting existing reasoning datasets into an iterative format for training models under this paradigm.

%% file: sec/part5_prompt.tex
\section{Input Prompts-based Efficient Reasoning}
From the perspective of input prompts and questions, these works focus on enforcing length constraints or routing LLMs based on the characteristics of input prompts to enable concise and efficient reasoning.

\subsection{Prompt-guided Efficient Reasoning}
\label{sec:prompts}

Prompt-guided efficient reasoning \textit{explicitly instructs LLMs to generate fewer reasoning steps}, can be a straightforward and highly effective method for improving the efficiency of reasoning models. As shown in Table~\ref{tab:prompts}, different methods propose different prompts to ensure concise reasoning outputs from the model.

\insightbox{The key question is: Which prompts can accurately control the reasoning length of LLMs?}

\begin{table}[t]
\centering
\small
\setlength{\tabcolsep}{4pt}
\renewcommand{\arraystretch}{1.2}
\caption{A summary of prompts used with reasoning models to generate concise reasoning outputs. For further details, refer to Section~\ref{sec:prompts}.}
\begin{tabularx}{\textwidth}{lX}
\toprule
\textbf{Method} & \textbf{Prompt} \\
\midrule
TALE-EP \cite{han2024token} & \textbf{Budget Estimation}: (...) Task: Analyze the given question and estimate the minimum number of tokens required to generate a complete and accurate response. Please give the response by strictly following this format: [[budget]], for example, Budget: [[12]].\\
& \textbf{Token-budget-aware CoT}: Please answer the above question. Let's think step by step and use less than \texttt{<Token-Budget>} tokens. \\
\midrule
CoD \cite{xu2025chain} & Think step by step, but only keep a minimum draft for each thinking step, with 5 words at most. Return the answer at the end of the response after a separator \texttt{\#\#\#\#}. \\
\midrule
CCoT \cite{renze2024benefits} & Be concise. \\
\midrule
Token Complexity~\cite{lee2025well} & \textbf{BulletPoints} (...) only use bullet points.\\
& \textbf{OnlyNumbers} (...) only use numbers or equations.\\
& \textbf{NoSpaces} (...) do not use any spaces or line breaks.\\
& \textbf{NoProperGrammar} (...) do not use proper grammar.\\
& \textbf{AbbreviateWords} (...) abbreviate words as much as possible.\\
& \textbf{WordLimit(k)} (...) use at most $k$ words. $k \in \{1, \dots, 100\}$\\
& \textbf{CharLimit(k)} (...) use at most $k$ letters. $k \in \{1, \dots, 500\}$\\
& \textbf{TokenLimit(k)} (...) use at most $k$ tokens. $k \in \{1, \dots, 500\}$\\
& \textbf{StepLimit(k)} (...) use at most $k$ steps. $k \in \{1, \dots, 5\}$\\
& \textbf{ChineseCoT} (...) Respond in Chinese\\
& \textbf{ChineseCoT(k)} (...) Use at most $k$ Chinese characters. $k \in \{1, \dots, 500\}$\\
\bottomrule
\end{tabularx}
\label{tab:prompts}
\end{table}

\paragraph{Enforcing Concise Reasoning via Varying Prompts.}
Token-Budget\cite{han2024token} proposes setting a token budget in prompts to reduce unnecessary reasoning tokens. To optimize efficiency while preserving accuracy, \cite{han2024token} introduced TALE-EP, a training-free, zero-shot method for budget estimation. TALE-EP first estimates a reasonable token budget by prompting the LLM itself. It then incorporates this estimate into a prompt that specifies the token constraint, guiding the LLM to generate a more token-efficient yet accurate response. This work is also categorized in Section ~\ref{sec:longshortdata} with further SFT.
CoD \cite{xu2025chain} observes that LLMs often generate excessively verbose reasoning steps, whereas humans typically record only the most essential insights. To enhance reasoning efficiency, they propose Chain-of-Draft prompting. Similar to CoT prompting, CoD encourages step-by-step reasoning but introduces policies to limit verbosity. For instance, their prompt instructs: ``\textit{Think step by step, but only keep a minimum draft for each thinking step, with at most five words.}'' They find that this approach preserves the necessary intermediate steps while maintaining accuracy, significantly reducing token usage.
\cite{lee2025well} systematically studies the relationship between reasoning length and model accuracy across various prompts with explicit compression instructions (e.g., ``\textit{use 10 words or less}''). Their analysis reveals a universal trade-off between reasoning length and accuracy, showing that different prompt-based compression strategies align on the same accuracy-compression curve. They hypothesize that each task has an intrinsic \textit{token complexity}, the minimum number of tokens required for successful problem-solving. By computing information-theoretic limits on the accuracy-compression trade-off, they found that existing prompt-based compression methods fall far short of these limits, indicating significant room for improvement.
\cite{renze2024benefits} introduced Concise Chain-of-Thought (CCoT) prompting, a technique that prompts LLMs to perform step-by-step reasoning while explicitly instructing them to ``\textit{be concise.}'' 
MARP \cite{chen2024unlocking} introduces modifying prompts to limit single-step computations, effectively refining the reasoning boundary. Further, they increase the per-step computation and decrease global planning steps.

\paragraph{Fine-tuning after Prompting.}
As noted in Section~\ref{tab:variable-length}, some approaches collect short CoT data using prompt-based methods, then apply SFT to develop an efficient reasoning model \cite{han2024token}. Beyond performing direct prompt-based reasoning, these fine-tuned models often deliver more promising performance when tackling complex reasoning challenges.

\subsection{Prompts Attribute-Driven Reasoning Routing}
\label{sec:routing}

User-provided prompts can range from easy to difficult tasks. Routing strategies for efficient reasoning dynamically determine how language models handle queries based on their complexity and uncertainty. Ideally, \textit{reasoning models can automatically assign simpler queries to faster but less reasoning-capable LLMs, while directing more complicated queries to slower but stronger reasoning LLMs.}

\insightbox{The key question is: What criterion should be used to determine the attributes (e.g., difficulty) of prompts?}

\paragraph{Unknown Criteria.} Anthropic releases Claude 3.7 Sonnet~\cite{anthropic_claude_sonnet}, notable for being the first hybrid reasoning model. Claude 3.7 Sonnet was developed through RL, enabling it to allocate more time to complex reasoning tasks that require deeper analysis, ultimately producing better results. The model offers two response modes: quick answers or step-by-step thinking. Users can leverage API to manage the amount of time the model spends thinking. Although the specifics of the routing criterion remain unclear, Claude 3.7 Sonnet represents the first hybrid reasoning model, setting a foundation for subsequent routing-based large reasoning models.

\paragraph{Training a Classifier.} RouteLLM~\cite{ong2024routellm} trains a query router to dispatch incoming queries to suitable LLMs based on complexity. The authors utilize a substantial amount of preference data collected from Chatbot Arena as training data, enabling effective routing decisions for question-answering and reasoning tasks. Consequently, simpler queries are directed to low-latency LLMs, while complex queries are assigned to higher-latency, more powerful LLMs, significantly accelerating overall reasoning efficiency. Sketch-of-Thought (SoT) \cite{aytes2025sketch} leverages routing and prompting to minimize token usage during reasoning. A lightweight DistilBERT-based router dynamically selects the most suitable paradigm based on the characteristics of the questions. Inspired by cognitive science, SoT employs three distinct paradigms: \textit{Conceptual Chaining}, which connects ideas with minimal verbalization; \textit{Chunked Symbolism}, which structures mathematical reasoning into concise symbolic representations; and \textit{Expert Lexicons}, which adopts domain-specific shorthand used by experts. 

\paragraph{Uncertainty.} Besides relying on additional routers, Self-Ref~\cite{chuang2025learningroutellmsconfidence} enables LLMs to autonomously decide when to route by extracting intrinsic uncertainty scores as self-routing indicators. Specifically, they fine-tune uncertainty-specialized tokens within the LLMs to align uncertainty predictions with prediction correctness in both question-answering and reasoning tasks. This ensures that only uncertain or incorrect outputs trigger routing to more capable LLMs, which decreases the latency of LLM inference. Confident or Seek Stronger~\cite{chuang2025confidentseekstrongerexploring} aims to provide calibrated data for predicting and initializing routing strategies in both LLM question-answering and reasoning tasks without requiring access to user queries. This approach enables more efficient and reliable decision-making in determining whether an LLM should confidently generate an answer or escort the query to a stronger model, ultimately improving reasoning efficiency from a query-level perspective in online LLM service scenarios.

%% file: sec/part6_lessdata.tex
\section{Reasoning Abilities via Efficient Training Data and Model Compression}

\subsection{Training Reasoning Models with Less Data}
\label{sec:lessdata}

Improving the efficiency of reasoning models requires optimizing not just the model architecture but also the data used for training. Recent work has shown that carefully selecting, structuring, and leveraging training data can significantly reduce data requirements while maintaining or even improving reasoning performance. Although all approaches focus on efficient data selection, they vary in defining and utilizing efficiency.

\insightbox{The key question is: How to construct less but high-quality training data?}

\paragraph{\textbf{Minimal but High-Impact Data Selection.}} LIMO \cite{ye2025limoreasoning} challenges the conventional belief that complex reasoning tasks require extensive training data. They introduce LIMO, a framework that elicits sophisticated reasoning abilities using minimal but precisely curated examples. By choosing high-quality questions based on \textit{Level of difficulty, Generality, and Knowledge Diversity} and high-quality solutions based on \textit{Optimal Structural Organization, Effective Cognitive Scaffolding, and Rigorous Verification}, with only 817 carefully selected training samples, LIMO can outperform previous models that utilized over 100,000 examples. 
s1 \cite{muennighoff2025s1simpletesttimescaling} focuses on enhancing reasoning performance by controlling test-time computational resources. They curate a compact dataset based on \textit{Quality}, \textit{Difficulty} and \textit{Diversity}, s1K, comprising 1,000 high-quality questions paired with reasoning traces. Through supervised fine-tuning on this dataset and implementing ``budget forcing'', which regulates the reasoning duration during inference, s1-32B exceeds OpenAI o1-preview on MATH and AIME24, demonstrating that strategic test time scaling can effectively enhance reasoning capabilities without extensive training data.

\paragraph{\textbf{Self-Verification as a Data-Efficient Training Signal.}} S$^2$R~\cite{ma2025s2rteachingllmsselfverify} infuse LLMs with self-verification and self-correction abilities through RL. Initially, models are fine-tuned on a curated dataset to establish these capabilities. Subsequently, RL both at the outcome level and the process level is employed to enhance these skills further. With only 3,100 initialization samples, their fine-tuned models consistently improve the performance on reasoning tasks among all base models. S$^2$R fine-tuned Qwen2.5-Math-7B can outperform models trained on comparable amounts of long CoT distilled data on the MATH500 and GSM8K.

%% file: sec/part8_smallmodel.tex
\subsection{Reasoning Capabilities of Small Language Models via Distillation and Model Compression}
\label{sec:compression}

LLMs have demonstrated remarkable reasoning capabilities across various complex tasks, benefiting from their extensive training on diverse datasets. However, their substantial computational and memory demands pose challenges for deployment in resource-constrained environments, such as edge devices, mobile applications, and real-time systems. In scenarios where efficiency, cost, or latency is a primary concern, Small Language Models (SLMs) offer a viable alternative. The ability of SLMs to retain strong reasoning capabilities while operating under strict resource constraints is crucial for expanding the accessibility and practicality of AI-powered reasoning systems. To achieve this, two main categories of approach are explored: Distillation and Model Compression.

\insightbox{The key question is: How do small language models perform on reasoning tasks? What impact does model compression (e.g., quantization) have on their reasoning abilities?}

\paragraph{\textbf{Distillation.}}

Distillation is a crucial technique for transferring the reasoning capabilities of LLMs to SLMs while maintaining efficiency. However, \cite{li2025small} finds a phenomenon named \textit{Small Model Learnability Gap}, which highlights the challenges of distilling complex reasoning processes from large model to small model, showing that SLMs struggle to emulate the reasoning depth of their larger counterparts. To address this, various approaches have been proposed. Both \cite{li2025small} and \cite{chenglin2024mixed} explored mixed distillation, with \cite{li2025small} blending long and short CoT reasoning examples, while \cite{chenglin2024mixed} combined CoT and PoT (Program of Thought) to improve the effectiveness of knowledge distillation from LLMs to SLMs on specific tasks. In comparison, \cite{feng2024teaching} introduced counterfactual distillation, augmenting the training set by masking causal features in the original question, prompting the LLM to complete the masked text, and generating multi-view CoT (positive and negative views) of each data for enhancing the effectiveness of knowledge distillation. 
In addition, \cite{zhu2024improving} developed a feedback-driven distillation technique that iteratively refines distillation datasets. They first prompt an LLM to generate an initial distillation dataset, then expand it by creating diverse and complex questions from existing ones, and finally, this enriched dataset is used to fine-tune SLMs. Another strategy, proposed by \cite{zhao2024probe}, incorporates probing and retrieval mechanisms into the distillation pipeline. It trains two complementary distilled SLMs, a probing model and a reasoning model, where the probing model retrieves relevant knowledge, which the reasoning model then uses to construct a step-by-step rationale for the answer. \cite{chen2024distilling} introduced adaptive thinking during distillation, allowing the models to dynamically adjust reasoning strategies based on the complexity of the task. 
Furthermore, \cite{liao2025skintern} proposed SKIntern, a framework that internalizes symbolic knowledge into SLM to improve CoT reasoning quality and efficiency, while \cite{zhang2024small} introduces SCORE, a pipeline that generates self-correction data from SLMs and fine-tunes the model to function as a self-correcting reasoner. These diverse distillation techniques demonstrate that efficiently transferring reasoning capabilities from LLMs to SLMs requires not only reducing the model size but also carefully and strategically structuring the knowledge transfer process to preserve logical depth and generalization.

\paragraph{\textbf{Pruning and Quantization.}}
Beyond directly distilling knowledge from LLMs to SLMs, an alternative approach involves compressing an LLM into an SLM using techniques such as quantization and pruning. \cite{srivastava2025towards} conducted a comprehensive study analyzing the impact of various model compression techniques on reasoning ability. Their findings reveal that \textit{quantization, which reduces model precision to lower-bit representations, preserves reasoning performance remarkably well}, allowing SLMs to maintain logical coherence and problem-solving capabilities while significantly reducing memory and computational costs. 

In contrast, \textit{pruning, which removes specific weights or neurons in the model based on their importance, leads to severe degradation in reasoning quality}, disrupting the model's ability to follow multi-step logical processes. This suggests that compression-based approaches are more effective than training SLMs from scratch, as they allow models to retain reasoning structures inherited from LLMs. However, a critical challenge remains: SLMs often struggle with the instruction following, indicating that compression alone is insufficient. Additional fine-tuning or adaptation methods may be required to align compressed models with user intent and ensure they can effectively interpret and execute complex reasoning tasks.

%% file: sec/part7_evaluation.tex
\section{Evaluation and Benchmark}
\label{sec:evaluation}

Recent research has introduced innovative benchmarks and evaluation frameworks to systematically assess the reasoning capabilities of LLMs. As LLMs continue to advance in their ability to perform complex reasoning tasks, the need for rigorous, standardized evaluation metrics and frameworks has become increasingly important. 

\paragraph{\textbf{Inference-time Computation.}} \cite{parashar2025inferencetimecomputationsllmreasoning} develops Sys2Bench, which is a comprehensive suite designed to evaluate LLMs across five reasoning categories, including arithmetic, logical, commonsense, algorithmic, and planning tasks. This benchmark comprises eleven diverse datasets, covering various reasoning tasks. It includes GSM8K and AQuA for arithmetic problems, StrategyQA and HotPotQA for commonsense reasoning, ProntoQA for logical reasoning, Game of 24 and Bin Packing for algorithmic tasks, and BlocksWorld, Rubik’s Cube, TripPlan, and Calendar Plan for planning tasks. The study revealed that scaling inference-time computation alone has limitations, as no single technique consistently excels across all reasoning tasks, and this emphasizes the need for diverse approaches to enhance LLM reasoning capabilities. Bag of Tricks \cite{liu2025bag} examines how various commonly used strategies affect the reasoning capabilities of LLMs. Furthermore, it benchmarks multiple inference-time computation methods within a predefined budget, enabling controlled token usage through a flexible N-sample strategy. \cite{liu20251bllmsurpass405b} investigates the impact of Test-Time Scaling (TTS) strategies on LLM performance, focusing on how policy models, process reward models, and problem difficulty influence TTS effectiveness. Their findings indicate that compute-optimal TTS strategies are highly dependent on these factors. The paper finds that, with appropriate TTS strategies, smaller models (e.g., a 1B parameter LLM) are able to outperform significantly larger models (e.g., a 405B parameter LLM) on complex reasoning tasks like MATH-500, and this underscores the importance of tailored TTS approaches in evaluating and enhancing LLM reasoning. 

\paragraph{\textbf{Evaluating Overthinking.}} \cite{cuadron2025dangeroverthinkingexaminingreasoningaction} introduces a framework to systematically analyze the "overthinking" in LLMs, where models favor extended internal reasoning over necessary environmental interactions. By examining 4,018 trajectories in agentic tasks, the study identified patterns such as Analysis Paralysis, Rogue Actions, and Premature Disengagement. \cite{cuadron2025dangeroverthinkingexaminingreasoningaction} also proposed a novel ``overthinking score'' and showed a strong correlation between higher scores and decreased task performance. Mitigation strategies such as selecting solutions with lower overthinking scores can improve performance by 30\% and at the same time reduce computational overhead by 43\%. \cite{liu2024mind} uses six comprehensive tasks from the overthinking literature to guide the design of evaluation benchmarks for testing CoT failures in LLMs. It reveals that, in many cases where humans tend to fail due to excessive deliberation, LLMs employing CoT reasoning exhibit similar failure patterns. S1-Bench~\cite{zhang2025s1} evaluates large reasoning models on straightforward tasks aligned with System 1 thinking, focusing on intuitive and fast reasoning rather than the more complex, deliberative processes associated with System 2.

\paragraph{\textbf{Effect of Long CoT Reasoning.}} \cite{yeo2025demystifying} provides a comprehensive analysis of the mechanism underlying long CoT reasoning. In addition to presenting several key insights, they propose a reward design to enhance the stability of reasoning ability during training and reduce the CoT length, which is also shown in Section~\ref{sec:rl}. \cite{jin2024impact} reveals a strong relationship between the length of the reasoning chain and the effectiveness of model outputs. Models tend to perform better with extended reasoning steps, suggesting the CoT length is more crucial than accuracy for effective problem-solving. CriticalThinking~\cite{lee2025criticalthinkingkindscomplexity} investigates the optimal reasoning length of LLMs based on using deterministic finite automata (DFAs). MiP-Overthinking~\cite{fan2025missing} discovers that when the reasoning models get questions missing key information, they tend to keep ``thinking'' over and over instead of admitting they fail to answer. By studying this behavior, they show that current training methods push models to generate long, repetitive reasoning steps rather than recognize unsolvable problems and stop early.

\paragraph{\textbf{Effect of Compression on Reasoning Models.}} CompressionReasoning~\cite{zhang2025reasoning} benchmarks compression techniques including quantization, distillation, and pruning, on reasoning tasks. The results indicate that parameter count has a greater impact on knowledge retention than on reasoning ability, and that generating shorter outputs generally leads to improved performance. QuantRM~\cite{liu2025quantizationhurtsreasoningempirical} provides a benchmark evaluating weight quantization, KV‑cache quantization, and activation quantization across various algorithms and bit‑width configurations.

%% file: sec/part11_application.tex
\section{Applications and Discussion}

\subsection{Applications}

\paragraph{Autonomic Driving.} Efficient reasoning LLMs are able to greatly improve autonomic driving \cite{cui2024survey, xing2025openemma, xing2024autotrust, xing2025can} by helping them understand large amounts of sensor data in a human-like way. They make the cars better at making decisions, so the vehicles can plan for difficult driving situations and react quickly when unexpected events occur. By combining information from cameras, LiDAR, radar, and other sensors, these models help cars drive more safely, choose better routes, and assess risks as they happen. Moreover, because they can explain why they make certain decisions, both passengers and regulators feel more confident in the technology, and the cars can interact more smoothly with smart road systems.

\paragraph{Embodied AI.} Efficient reasoning LLMs make embodied AI \cite{duan2022survey} much smarter by helping robots and smart devices understand and react to the world around them. These models process lots of data from cameras, sensors, and other inputs in a way that resembles human thinking. This deep understanding means that a robot can quickly decide the best way to move, handle unexpected changes, and interact safely with people. For example, in a busy factory or a home setting, a robot using these models can navigate obstacles, adjust to new situations, and even explain its actions in simple terms. Altogether, efficient reasoning LLMs boost the reliability, safety, and usefulness of embodied AI systems in daily environments.

\paragraph{Healthcare.} Efficient reasoning LLMs would improve healthcare \cite{he2023survey} by helping doctors and researchers work with large amounts of medical data more easily. They can quickly analyze patient records, test results, and medical research to spot important trends and patterns that might be hard to see otherwise. This support can lead to faster and more accurate diagnoses, better treatment recommendations, and fewer mistakes. In addition, these models can break down complex medical information into plain language, making it easier for both medical professionals and patients to understand. Generally, efficient reasoning LLMs make healthcare processes smoother and more reliable, leading to better care and outcomes for patients.

\paragraph{Recommender System.} Efficient reasoning LLMs can greatly enhance recommender systems by enabling more accurate, personalized, and context-aware suggestions across various domains such as e-commerce, entertainment, and education. These models can reason over diverse and dynamic user behavior, preferences, and historical interactions to uncover subtle patterns and relationships that traditional models might overlook. By efficiently processing complex input data, LLMs can generate high-quality recommendations with fewer computational resources. For instance, in an online shopping platform, an efficient reasoning model can anticipate evolving user interests, adapt to seasonal trends, and explain recommendations in a user-friendly way. Overall, efficient reasoning LLMs improve the scalability, transparency, and responsiveness of recommender systems, leading to better user satisfaction and engagement. ReaRec \cite{tang2025think} proposes a latent reasoning framework for recommender systems. Inspired by the think-before-action paradigm in LLMs, ReaRec enables implicit multi-step reasoning at inference time, significantly improving performance, particularly for long-tail users and items. 

\subsection{Discussion}

\textbf{Improving Reasoning Ability.} From another perspective on efficiency, improving reasoning performance is an important topic \cite{chen2025inner, sui2025meta}. To prioritize promising avenues by discarding ineffective strategies early, Meta-Reasoner \cite{sui2025meta} leverages contextual multi-armed bandits for evaluating reasoning progress and selecting the optimal strategy. In each round, the LLM produces a new reasoning step, and the meta-reasoner evaluates its output and generates a progress report, the meta-reasoner uses contextual multi-arm bandit to choose the best guidance strategy for the reasoning step. ITT \cite{chen2025inner} treats each transformer layer as a step in an internal thinking process. By dynamically allocating extra processing to difficult tokens through adaptive routing, ITT enables smaller language models to achieve performance comparable to larger models while using fewer training resources. SyzygyoT~\cite{li2025syzygy} introduces Minimal Free Resolution (MFR), which is inspired by algebraic geometry, to break down complex tasks into logically complete and minimal subproblems. This decomposition enhances the structural efficiency of CoT reasoning, enabling more precise and efficient problem-solving.
 
\textbf{Safety of Efficient Reasoning.} Safety and efficiency in LLMs often pull in opposite directions, as optimizing one always leads to the performance degradation of the other. When enhancing safety, such as filtering harmful content, mitigating adversarial attacks, and enabling self-correction, the reasoning model typically requires additional computational resources and longer reasoning sequences, leading to increased inference costs and slower response times. Conversely, prioritizing efficiency by minimizing token usage and computational overhead may reduce the reasoning ability to self-reflect, verify its outputs, or defend against adversarial manipulations. This trade-off reflects the well-known principle that there is no ``free lunch'', making it crucial to strike a careful balance between safety and efficiency. \cite{kuo2025h} investigates the robustness of safety checks in large CoT reasoning models, revealing severe security flaws in commercial systems. They introduce the malicious-educator benchmark and demonstrate that with their hijacking Chain-of-Thought (H-CoT) attack, models can drastically reduce their refusal rates, leading to the generation of harmful content. \cite{li2025output} investigates the safety of long reasoning models. It is observed that while longer outputs enable self-correction and enhance safety, some attack strategies exploit extended generations. They propose a dynamic output length control via an RL-based method to maintain both reasoning quality and security. Balancing safety and efficiency in long reasoning models remains a challenging yet crucial area of investigation. \cite{kumar2025overthink} proposes an indirect prompt injection attack targeting reasoning LLMs applied to untrusted data sources and substantially degrades reasoning efficiency. SafeMLRM~\cite{fang2025safemlrm} provides a safety analysis of multi-modal large reasoning models (MLRMs).

\textbf{Efficient LLMs for Agentic AI.} Efficient reasoning is essential to the advancement of agentic AI systems~\cite{liu2025advances}, as it directly influences their decision-making speed, resource utilization, and overall effectiveness in real-world applications. Recent research efforts have extensively investigated methods for improving agent efficiency by optimizing internal reasoning processes. Notable approaches include merging multiple planning trees to reduce computational redundancy~\cite{hu2023tree}, as well as consolidating and structuring memory representations to enhance efficiency and adaptability in dynamic environments~\cite{hu2024hiagent}. CoA~\cite{pan2024chain} enhances the capability of LLMs in managing complex tasks, especially in scenarios demanding real-time or domain-specific knowledge, with an efficient verification module leveraging our MRFS framework to refine LLM-generated responses using retrieved information. These innovations collectively contribute toward the development of more responsive, scalable, and practical AI agents capable of operating effectively under constrained computational resources. DOWN~\cite{eo2025debatenecessaryadaptivemultiagent} proposes an adaptive multi‑agent debate framework that only triggers debate when the initial confidence of an agent is low, then uses the confidence‑weighted inputs of participating agents to collaboratively refine the final answer.

\textbf{RL vs. SFT, which is better?} When comparing RL (Section \ref{sec:rl}) and SFT (Section \ref{sec:longshortdata}) for creating efficient reasoning language models, the answer is unclear as each method has its own strengths. RL allows a model to learn by trial and error, rewarding it for satisfactory decisions, which can assist it find creative ways to solve problems in new situations. However, this approach can sometimes be unpredictable and require a lot of training. On the other hand, SFT teaches the model using carefully chosen efficient CoT examples constructed by either humans or models, leading to more consistent behavior and easier control. Yet, SFT might struggle when faced with challenges that are not covered in its training data. In practice, combining both methods might be a promising direction and potentially works best because it harnesses the creativity of RL and the reliability of SFT, resulting in a model that is both adaptable and stable.

%% file: sec/5_conclusion.tex
\section{Conclusion}

This paper provides the first structured survey of efficient reasoning in LLMs, categorizing existing approaches into three areas: model-based, reasoning output-based, and input prompts-based methods. Additionally, it discusses efficient data utilization, reasoning capabilities of smaller models, evaluation techniques, and benchmarking, accompanied by a continuously updated public repository to support future research. Crucially, efficient reasoning approaches offer significant practical benefits across various domains: reducing computational costs in healthcare diagnostics, enhancing real-time decision-making and safety in autonomous driving, boosting the reliability and usefulness of embodied AI systems, and enabling quicker, more profitable responses in financial algorithmic trading and risk assessment. These advancements highlight the broad economic and societal value of efficient reasoning in LLMs.